\documentclass[letterpaper]{article} 
\usepackage{aaai2027}  
\nocopyright
\usepackage[hyphens]{url}  
\usepackage{graphicx} 
\usepackage{natbib}  
\usepackage{caption} 
\usepackage[utf8]{inputenc} 
\DeclareUnicodeCharacter{266B}{\ding{14}} 
\usepackage{algorithm}
\usepackage{algorithmic}
\usepackage{booktabs}
\usepackage[table]{xcolor}
\usepackage[most]{tcolorbox}
\usepackage{multirow}
\usepackage{amsmath}
\usepackage{amssymb}
\usepackage{makecell}
\usepackage{pifont}
\definecolor{avgblue}{RGB}{220,230,245}

\usepackage{newfloat}
\usepackage{listings}
\DeclareCaptionStyle{ruled}{labelfont=normalfont,labelsep=colon,strut=off} 
\floatstyle{ruled}
\newfloat{listing}{tb}{lst}{}
\floatname{listing}{Listing}

\usepackage{booktabs}
\usepackage{amsmath}
\usepackage{amssymb}

\newcommand{\method}{\textsc{LoongReflect}}
\newcommand{\reflect}{\texttt{<reflect>}}
\newcommand{\backtrack}{\texttt{<backtrack>}}

\newcommand{\answer}{\texttt{<answer>}}

\title{\method: Boosting Long-Horizon Reflection in Search Agents via Global Perspective Distillation}
\author{
    Zhixin Zhang\equalcontrib\textsuperscript{\rm 1,\rm 2,\rm 3},
    Xinke Jiang\equalcontrib\textsuperscript{\rm 1,\rm 2,\rm 3},
    Zhibang Yang\equalcontrib\textsuperscript{\rm 1,\rm 2},
    Weixuan Xu\textsuperscript{\rm 1,\rm 2,\rm 3},\\
    Guohong Qiu\textsuperscript{\rm 1,\rm 2,\rm 3},
    Xu Chu\corresponding\textsuperscript{\rm 1,\rm 3,\rm 4},
    Junfeng Zhao\corresponding\textsuperscript{\rm 1,\rm 3},
    Yasha Wang\corresponding\textsuperscript{\rm 2,\rm 5}
}
\affiliations{
    \textsuperscript{\rm 1}School of Computer Science, Peking University, Beijing, China\\
    \textsuperscript{\rm 2}National Engineering Research Center for Software Engineering, Peking University, Beijing, China\\
    \textsuperscript{\rm 3}Key Laboratory of High Confidence Software Technologies, Ministry of Education, Beijing, China\\
    \textsuperscript{\rm 4}Center on Frontiers of Computing Studies, Peking University, Beijing, China\\
    \textsuperscript{\rm 5}Peking University Information Technology Institute (Tianjin Binhai), Tianjin, China\\
    \{zhixinzhang25,xinkejiang,yangzb\}@stu.pku.edu.cn,\\
    \{chu\_xu, zhaojf, wangyasha\}@pku.edu.cn
}

\begin{document}

\maketitle

\begin{abstract}
Large language model agents increasingly rely on long-horizon reasoning to solve complex tasks involving planning, tool use, and memory.
A critical capability in such settings is~\textbf{reflection}: assessing trajectory progress, identifying missing evidence and unreliable intermediate states, and deciding whether to continue, revise, or abandon the current branch. 
Learning effective reflection, however, is challenging because reflection is performed locally within the current branch, whereas its utility can only be determined by its contribution to the final trajectory outcome. This local--global mismatch makes outcome-based reinforcement learning provide only local, sparse and delayed supervision for reflective decisions.
To solve these, we propose~\textbf{\method}, a training framework that formulates reflection as a~\textit{memory-control policy}. The agent operates over a reversible trajectory tree using explicit~\textbf{\reflect{}} and~\textbf{\backtrack{}} actions. Reflection consolidates verified facts, missing evidence, and branch-specific risks into working memory, while backtracking removes an unreliable branch from the active context and preserves a concise corrective lesson. To learn this policy,~\method~combines two complementary signals through a look-ahead, extragradient-style coordination mechanism. A~\textbf{\textit{fast channel}} distills globally informed reflective behavior from a privileged teacher, with supervision restricted to reflection and backtracking tokens. A~\textbf{\textit{slow channel}} optimizes complete trajectories using outcome-based GRPO, aligning local control decisions with final task success. Experiments on multi-hop retrieval-augmented generation and mathematical reasoning benchmarks demonstrate consistent improvements over outcome-only reinforcement learning and self-distillation baselines.
\end{abstract}

\section{Introduction}

Large language models (LLMs) have emerged as the core policy components of agents that reason, retrieve information, and maintain intermediate memory \citep{yao2022react,schick2023toolformer,asai2024self,Agentic_RAG_R1}. As these systems are increasingly deployed on open-ended, multi-step tasks, long-horizon reasoning becomes critical. In such settings, an agent must do more than generate the next action: it must also continually assess whether its current trajectory is making progress, whether the available evidence is sufficient, and whether an earlier branch should be revised or abandoned, which is referred to as \textbf{reflection}.

\begin{figure}[t]
\centering
\includegraphics[width=\linewidth]{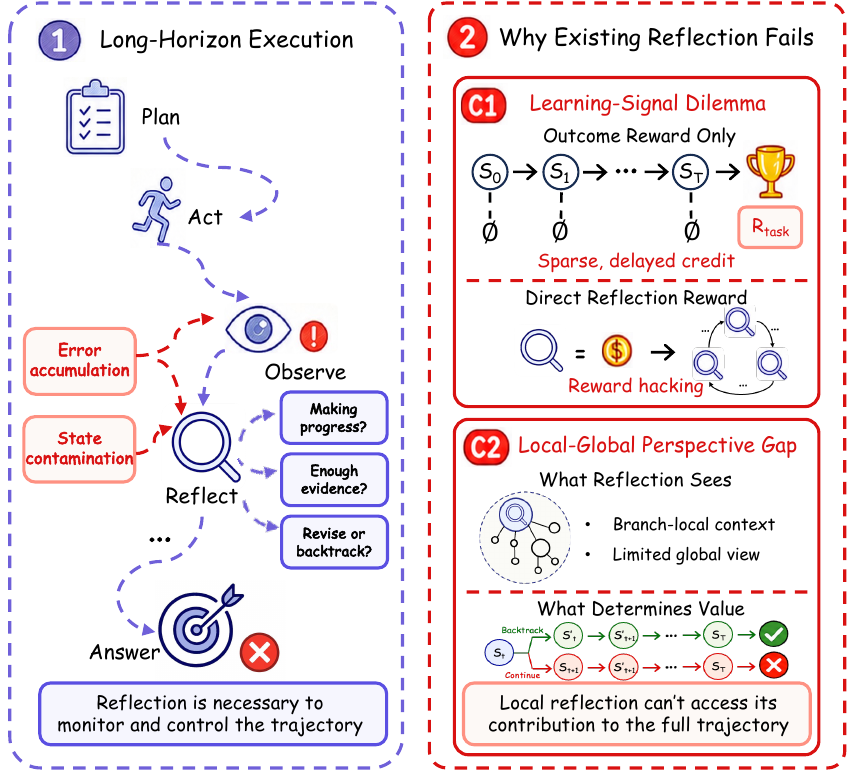}
\caption{Reflective decisions are made from branch-local context, whereas their value is determined by their contribution to the final trajectory outcome. This local--global mismatch leaves outcome-based learning with local, sparse and delayed supervision.}
\label{fig:intro}
\end{figure}

\textbf{Reflection} is not merely additional reasoning text; it is a control process over the trajectory itself. An effective reflection step should consolidate verified facts, identify missing evidence, diagnose branch-specific risks, and decide whether the current branch should be continued, revised, or abandoned~\citep{jiang2025tc,Agentic_RAG_R1,zhang2026stackplanner}. This capability is particularly important over long horizons, where localized errors---such as irrelevant retrievals, spurious entity associations, or stale memory updates---can enter the active context and influence many subsequent decisions. Without an explicit mechanism for isolating and correcting such errors, state contamination compounds as the trajectory grows, becoming a major obstacle to reliable agent behavior~\citep{zhu2025llm,jiang2025tc,Agentic_RAG_R1,zhang2026stackplanner}.

Despite its importance, learning effective reflection faces two fundamental challenges, as illustrated in Figure~\ref{fig:intro}. 
\textbf{\ding{182}  \textit{(\textit{C1}) Learning-signal dilemma.}} The value of a reflective decision is mediated by many subsequent actions and is revealed only through the final task outcome. Outcome-based reinforcement learning therefore provides sparse, delayed, and weakly attributable supervision for reflection. Yet assigning reflection an explicit intermediate reward is equally problematic and prone to reward hacking, as it may encourage excessive or superficial reflection without improving task success.
\textbf{\ding{183}  \textit{(C2) Local--global perspective gap.}} Reflection is performed from the context of the current branch, whereas its true value depends on how that branch contributes to the complete trajectory. A local reflector cannot directly observe whether continuing, revising, or abandoning the branch will ultimately improve the outcome. Together, these challenges create a fundamental mismatch: \textbf{effective reflection must be learned from dense local guidance, yet evaluated and calibrated from a global trajectory perspective}.

Existing approaches address only part of this problem. \ding{182} One line of work improves intermediate judgment through step-level verification or action evaluation~\citep{lightman2024let,wang2024math,liu2026trust,liu2026agentic,guo2025deepseek}. \ding{183} Another provides denser learning signals through privileged feedback or error-localized supervision~\citep{agarwal2024policy,zhao2026self,zhao2026rosd,li2026localizing,huang2026learning}. 
However, these approaches either treat reflection primarily as \textbf{local judgment} or rely on \textbf{misaligned supervision signals} that are not fully consistent with long-horizon outcomes. They therefore fall short of learning reflection as trajectory-level control for long-horizon agents.

To address \textbf{\textit{C1\&C2}}, we propose \textbf{\method}, a training framework that formulates reflection as a \textbf{memory-control policy}, designed for long-horizon setting. Rather than treating reflection as unconstrained verbal feedback, \method turns it into explicit, structured decisions over what the agent should \textbf{retain, revise, or discard from its reasoning memory}. Specifically, the agent operates over a \textbf{reversible trajectory tree} with an explicit active path and two control actions: \textbf{\reflect{}} and \textbf{\backtrack{}}. The \reflect{} action consolidates verified facts, missing evidence, and branch-specific risks into working memory. When the current branch is deemed unreliable, the \backtrack{} action removes its contaminated suffix from the active context, restores a validated state, and preserves a concise corrective lesson for subsequent decisions.

We train this memory-control policy through two complementary learning channels designed to resolve the two challenges jointly. \ding{182} A \textbf{fast channel} distills reflective behavior from a privileged teacher, with supervision restricted to reflection and backtracking tokens. This token-level supervision provides the \textbf{dense} learning signal missing from outcome-based reinforcement learning, thereby addressing \textbf{\textit{C1}}. Moreover, because the teacher observes the trajectory tree and terminal outcome, it can evaluate a local branch from a \textbf{global} perspective, thereby addressing \textbf{\textit{C2}}. To prevent answer imitation, the teacher produces answer-masked feedback that focuses exclusively on state diagnosis, missing evidence, branch risks, and control decisions. \ding{183} A \textbf{slow channel} applies outcome-based GRPO to complete trajectories, ensuring that the distilled reflective behavior remains aligned with final task success. We coordinate the two channels through a look-ahead, extragradient-style update, in which the slow objective evaluates and calibrates the update direction proposed by the fast channel before the combined update is committed.
Our contributions are as follows:
\begin{itemize}
    \item We formulate reflection as a \textbf{memory-control policy} for long-horizon agents and instantiate it through explicit \reflect{} and \backtrack{} actions over a reversible trajectory tree.
    
    \item We introduce a two-channel learning framework that combines answer-masked teacher distillation for dense, globally informed reflective supervision with outcome-based GRPO for trajectory-level alignment, coordinated through a look-ahead, extragradient-style update.
    
    \item \method\ consistently outperforms outcome-only RL and self-distillation baselines on multi-hop RAG and mathematical benchmarks, demonstrating the contributions of structured reflection, reversible backtracking, and two-channel optimization.
\end{itemize}
\section{Related Work}

\subsection{Long-Horizon Agents}

Large language models increasingly serve as the policy backbone of agents that interleave reasoning, retrieval, and tool use over extended interactions~\citep{yao2022react,schick2023toolformer,asai2024self}. Recent studies apply reinforcement learning to optimize long-horizon interaction and search policies directly from task outcomes~\citep{chen2025reinforcement,gao2025beyond}. Memory-oriented approaches further support extended execution by compressing interaction histories or explicitly manipulating intermediate states~\citep{jiang2025tc,zhang2026stackplanner}. For example, MEM1 jointly optimizes reasoning and compact memory, while AgenticRAG-R1 introduces stack-based memory with planning, summarization, and backtracking actions~\citep{zhou2025mem1,Agentic_RAG_R1}. However, these methods largely rely on task-level objectives or generic action-quality signals, offering limited supervision for identifying which local decision corrupted the execution state or whether a recovery action was effective. A complementary line of work provides denser corrective signals through on-policy distillation: OPSD distills a privileged self-teacher on trajectories generated by the current policy, whereas ROSD localizes supervision to the first erroneous span to preserve valid prefixes~\citep{zhao2026self,zhao2026rosd}. Nevertheless, these methods are primarily designed for linear reasoning traces and \textbf{provide limited support for reversible branch-level recovery or for aligning local corrections with delayed task outcomes}.

\subsection{Self-Reflection and Self-Correction}

Self-reflection and self-correction have been widely studied as inference-time mechanisms for iterative diagnosis and revision. Self-Refine repeatedly critiques and improves an initial response, while Reflexion stores verbal feedback from previous attempts as episodic experience~\citep{madaan2023self,shinn2023reflexion}. External tools can further ground reflective feedback and improve correction reliability~\citep{gou2024critic}. Nevertheless, intrinsic self-correction remains unreliable when models must evaluate their own generations without verifiable feedback~\citep{kamoi2024can}. Recent learning-based methods therefore introduce explicit supervision for reflection and intermediate decisions. RISE jointly trains solution generation and self-verification on current-policy outputs~\citep{liu2026trust}, while Agentic Critical Training learns to distinguish superior actions from suboptimal alternatives~\citep{liu2026agentic}. Process-supervised approaches further estimate intermediate action quality through step-level evaluation or automatically constructed reward models~\citep{xiong2024watch,deng2025agentpro}. Although these methods provide finer-grained supervision than terminal rewards, they mainly \textbf{target verification or action selection, with limited support for explicit state recovery and aligning local reflection with delayed long-horizon outcomes}.
\section{Method}
\label{sec:method}

\begin{figure*}[t]
    \centering
     \IfFileExists{figs/method_overview.pdf}{%
        \includegraphics[width=0.9\textwidth]{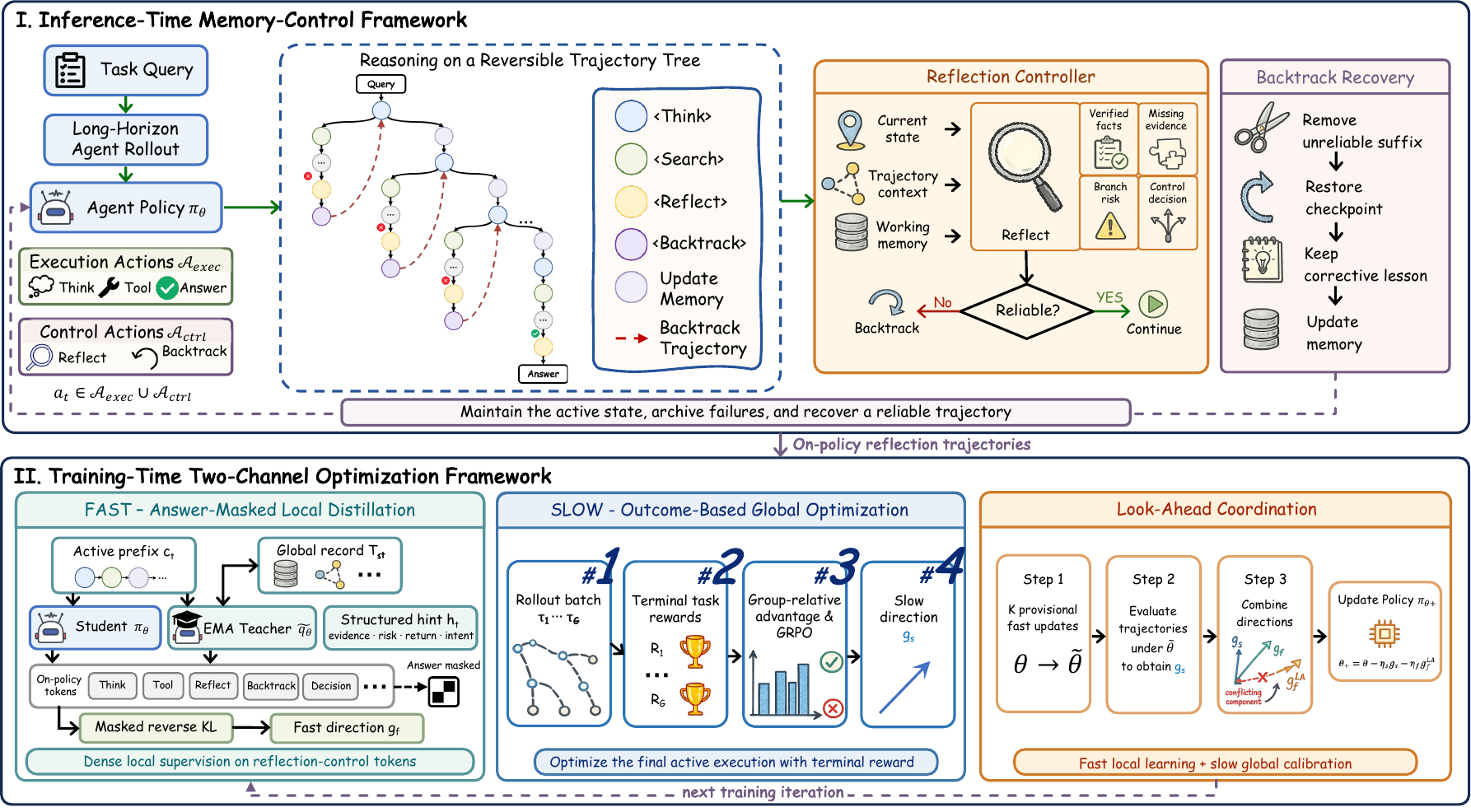}%
    }{%
        \fbox{\parbox[c][4.4cm][c]{0.96\textwidth}{
            \centering
            \textbf{Placeholder for the two-column method overview.}\\[3pt]
            Left: reflective interaction and reversible state transitions over a trajectory tree.\quad
            Right: answer-masked Fast reflection distillation, outcome-driven Slow optimization, and the Look-Ahead Update.
        }}%
    }
    \caption{Overview of the \method~ framework.}
    \label{fig:method_overview}
\end{figure*}

In this section, we present \method, as illustrated in Figure~\ref{fig:method_overview}. It consists of two components:
\ding{182} a memory-control framework that supports structured reflection through a reversible trajectory tree and explicit reflection actions; and
\ding{183} a two-channel optimization framework that combines dense local distillation with outcome-based global correction through look-ahead updates.

\subsection{Problem Formulation}\label{sec:formulation}

Given a task $x$, \method models a long-horizon agent as both an \emph{execution policy} and a \emph{memory-control policy}. In addition to reasoning, retrieval, and answer generation, the agent must determine which information accumulated during execution should be retained, revised, or discarded. We therefore equip the agent with a reversible trajectory state that can be explicitly modified through reflection and backtracking.
At step $t$, the agent state is defined as
\begin{equation}
\mathbf{z}_t
\;=\;
\bigl(x,\;\mathcal{T}_t,\;P_t,\;\mathbf{m}_t\bigr),
\label{eq:agent_state}
\end{equation}
where $\mathcal{T}_t$ denotes the accumulated trajectory tree, $P_t$ the active execution path, and $\mathbf{m}_t$ the working memory compressed from that path. Although the agent executes actions sequentially along $P_t$, its complete interaction history is not necessarily linear. When the agent backtracks to an earlier state and resumes execution, the newly generated suffix forms an alternative branch, while the abandoned suffix remains in $\mathcal{T}_t$ as an inactive branch. Repeated recovery therefore naturally induces a tree structure, with each root-to-node path representing a distinct execution branch.

Only the active path $P_t$ and its compressed memory $\mathbf{m}_t$ are exposed to subsequent generation. Inactive branches are excluded from the active context to prevent unreliable states from influencing future decisions, but remain in $\mathcal{T}_t$ for error diagnosis and reflective supervision. This separation enables the agent to recover from a corrupted branch without discarding the information needed to learn from it.

Conditioned on $\mathbf{z}_t$, the agent selects actions from both execution and control spaces:
\begin{equation}
a_t
\;\sim\;
\pi_{\boldsymbol{\theta}}
\bigl(\cdot \mid \mathbf{z}_t\bigr),
\qquad
a_t \in
\mathcal{A}_{\mathrm{exec}}
\cup
\mathcal{A}_{\mathrm{ctrl}},
\label{eq:action_policy}
\end{equation}
where $\mathcal{A}_{\mathrm{exec}}$ contains reasoning, tool-use, and answer actions, while $\mathcal{A}_{\mathrm{ctrl}}$ contains reflection and backtracking actions. Execution actions advance task completion, whereas control actions assess and regulate the validity of the active state.
The overall learning objective is to maximize the terminal task reward:
\begin{equation}
\max_{\boldsymbol{\theta}}
\;\;
\mathbb{E}_{\tau \sim
\pi_{\boldsymbol{\theta}}(\cdot \mid x)}
\bigl[R(\tau)\bigr],
\label{eq:task_objective}
\end{equation}
where $R(\tau)$ is determined by the final outcome of trajectory $\tau$. This objective alone, however, provides only sparse trajectory-level feedback for reflection decisions made at intermediate states. We therefore introduce a structured memory-control mechanism together with a two-channel optimization procedure, as described in the following sections.

\subsection{Reflection as Memory Control}\label{sec:reflection_memory_control}

To support reflection over intermediate states, \method represents agent memory as a reversible trajectory tree. At step $t$, the execution state consists of a trajectory tree $\mathcal{T}_t=(\mathcal{V}_t,\mathcal{E}_t)$, an active path $P_t$, compressed memory $\mathbf{m}_t$, and archived branches $\mathcal{B}_t$. The LLM context is serialized as
\begin{equation}
c_t=\mathrm{Serialize}\bigl(x,\,P_t,\,\mathbf{m}_t\bigr).
\end{equation}

Thus, the model accesses only the task, active path, and compressed memory, while $\mathcal{T}_t$ and $\mathcal{B}_t$ remain internal states for future diagnosis and recovery. This decouples current decision making from historical branches. Beyond task-solving actions, the agent may trigger reflection controls:
\begin{equation}
a_t^{\mathrm{ctrl}} \in \mathcal{A}_{\mathrm{ctrl}}
= \{\,\reflect{},\;\backtrack{}\,\}.
\end{equation}



These actions form the memory-control mechanism:
\begin{itemize}
\item \textbf{\reflect{}} diagnoses the active state by summarizing evidence, identifying missing information or faulty assumptions, and proposing the next control decision.

\item \textbf{\backtrack{}} executes recovery when reflection determines that the current state is unreliable, rolling the active trajectory back to a trustworthy prefix and removing
\end{itemize}
Concretely, \textbf{\reflect{}} produces a structured summary.
\begin{equation}
\begin{aligned}
\mathbf{r}_t &= \bigl(\,e_t^{\mathrm{ver}},\; q_t^{\mathrm{risk}},\;
j_t^{\mathrm{ret}},\; d_t^{\mathrm{ctrl}}\,\bigr),\\[2pt]
d_t^{\mathrm{ctrl}} &\in \{\,\mathrm{continue},\;\mathrm{backtrack}\,\}.
\end{aligned}
\end{equation}
where $e_t^{\mathrm{ver}}$, $q_t^{\mathrm{risk}}$, $j_t^{\mathrm{ret}}$, and $d_t^{\mathrm{ctrl}}$ denote evidence, risk, return point, and control intent, respectively.
The control intent determines whether to continue along the active path or invoke \textbf{\backtrack{}} for recovery, making reflection a state-control diagnosis rather than a generic critique. When reliable, the active path grows forward; otherwise, the system restores a reliable prefix and performs the transition
\begin{equation}
P_{t+1}=P_j \oplus u_{j:t},
\qquad
\mathcal{B}_{t+1}=\mathcal{B}_t\cup\{\,P_{j+1:t}\,\},
\end{equation}
where $P_j$ is the recovered active prefix, $P_{j+1:t}$ is the invalidated suffix removed from the active context, and $u_{j:t}$ is a compact corrective update distilled from the current reflection, such as the decisive contradiction, the falsified assumption, or the constraint that the next attempt must satisfy.

\subsection{Two-Channel Optimization for Reflection Learning}\label{sec:two_channel_optimization}

\paragraph{Fast Local Supervision.}

To provide dense supervision for reflection learning, \method introduces a fast channel over reflection control tokens. At step $t$, the student generates an active-state prefix with context $c_t=(x,P_t,\mathbf{m}_t)$. A teacher then constructs an answer-masked structured control label from the execution history:
\begin{equation}
\mathbf{h}_t=\mathcal{H}_{\phi}\bigl(x,\,\mathcal{T}_{\le t},\,P_t,\,\mathbf{m}_t\bigr),
\end{equation}
where $\mathcal{T}_{\le t}$ includes both the current active path and archived historical branches, and $\mathcal{H}_{\phi}$ denotes a privileged hint constructor implemented by either an auxiliary LLM or a rule-based feedback module. Its output shares the same schema as the reflect summary,
\begin{equation}
\mathbf{h}_t=\bigl(\,e_t^{\mathrm{ver}},\; q_t^{\mathrm{risk}},\; j_t^{\mathrm{ret}},\; d_t^{\mathrm{ctrl}}\,\bigr),
\end{equation}

but is constructed from privileged access to the global execution record while masking the final answer, so the fast channel supervises local diagnosis and recovery rather than answer generation. The student and teacher then evaluate the same continuation. Let $y_{t,k}$ denote the $k$-th generated token at step $t$. We define:
\begin{equation}
\begin{aligned}
\ell_{t,k} &= \log \pi_\theta\bigl(y_{t,k}\mid c_t,\,y_{t,<k}\bigr),\\[2pt]
\bar{\ell}_{t,k} &= \log q_{\bar\theta}\bigl(y_{t,k}\mid c_t,\,\mathbf{h}_t,\,y_{t,<k}\bigr).
\end{aligned}
\end{equation}
where $q_{\bar\theta}$ is an EMA teacher of the policy. The teacher and student share the same on-policy prefix and continuation; the only additional information available to the teacher is the structured hint $\mathbf{h}_t$. Since the fast channel supervises only reflection-related spans, we define the token mask
\begin{equation}
m_{t,k}^{\mathrm{ref}}
=
\mathbf{I}\;\bigl[\,y_{t,k}\in \mathrm{Span}(\reflect{},\backtrack{})\,\bigr], \notag
\end{equation}
with normalization factor
$Z=\sum_{t,k}m_{t,k}^{\mathrm{ref}}$.
Based on the teacher--student log-probability gap
$\delta_{t,k}=\ell_{t,k}-\bar{\ell}_{t,k}$, we optimize a masked reverse-KL objective (a k3-style unbiased estimator) restricted to the reflection span:
\begin{equation}
\mathcal{L}_{\mathrm{fast}}
=
\frac{1}{Z}
\sum_{t,k}
m_{t,k}^{\mathrm{ref}}\,
\min\!\bigl(\exp(-\delta_{t,k})-1+\delta_{t,k},\;c\bigr),
\end{equation}
where the clipping constant $c$ suppresses extreme token-ratio estimates. This objective is applied only to \textbf{\reflect{}} and \textbf{\backtrack{}} tokens, providing dense local supervision for state diagnosis and recovery decisions.

\paragraph{Slow Global Optimization.}



To calibrate the global utility of reflection, \method introduces a slow channel that optimizes reflection decisions based on terminal trajectory outcomes. Unlike the fast channel, which supervises local state diagnosis, the slow channel evaluates whether these decisions improve final task success. Specifically, for each task $x$, we sample complete trajectories $\tau_1,\tau_2,\ldots,\tau_G\sim\pi_\theta(\cdot\mid x)$ with terminal rewards $R_1,R_2,\ldots,R_G$, where $R_i$ is a task-level reward returned by the environment or an answer verifier. 
The slow channel thus evaluates reflection by trajectory-level success rather than the plausibility of individual control decisions. Given the trajectory group, we define the group-relative advantage as
$
\widehat{A}_i=
{\bigl(R_i-\operatorname{mean}_{g}(R_g)\bigl)}/
{\operatorname{std}_{g}(R_g)}.
$


This normalizes rewards within the sampled group and assigns credit based on relative trajectory quality. To ensure clear credit assignment, the slow objective is applied only to policy-generated tokens in the final active execution. 
Let $\xi_i$ denote the policy token sequence of trajectory $\tau_i$, including execution and reflection control tokens (\textbf{\reflect{}} and \textbf{\backtrack{}}), while excluding tool outputs, external observations, and controller updates. We then define the token ratio with respect to the old policy as
\begin{equation}
\rho_{i,k}(\theta)=
\frac{\pi_\theta(y_{i,k}\mid c_{i,k})}
{\pi_{\mathrm{old}}(y_{i,k}\mid c_{i,k})},
\end{equation}
where $c_{i,k}$ denotes the conditional prefix for the $k$-th token in trajectory $\tau_i$. Let $\bar{\rho}_{i,k}(\theta)=\operatorname{clip}(\rho_{i,k}(\theta),1-\varepsilon,1+\varepsilon)$. The slow channel is optimized by
\begin{equation}
\begin{aligned}
\mathcal{L}_{\mathrm{slow}}(\theta)
={}&-\frac{1}{G}\sum_{i=1}^{G}\frac{1}{|\xi_i|}
\sum_{k\in\xi_i}\min\!\left\{
\rho_{i,k}(\theta)\widehat A_i,
\bar{\rho}_{i,k}(\theta)\widehat A_i\right\}\\
&+\beta\,D_{\mathrm{KL}}
\bigl(\pi_\theta\,\|\,\pi_{\mathrm{ref}}\bigr). \notag
\end{aligned}
\end{equation}
Since $\widehat{A}_i$ is derived from terminal outcomes, slow channel rewards reflection only when it improves overall performance.
\paragraph{Look-Ahead Coordination.}


The fast and slow channels provide local and global signals, but their update directions may conflict. \method introduces look-ahead coordination, where slow channel calibrates fast update before optimization. Given $\theta$, we obtain a provisional fast policy $\widetilde{\theta}=\mathcal{U}_{\mathrm{fast}}^{K}(\theta)$ by applying $K$ inner fast-channel updates estimated direction:
\begin{equation}
g_f=(\theta-\widetilde{\theta})/\alpha,
\end{equation}
where $\alpha$ is accumulated inner step size. We then evaluate trajectories under $\widetilde{\theta}$ to derive slow-channel calibration direction
\begin{equation}
g_s=\nabla_{\widetilde{\theta}}\,\mathcal{L}_{\mathrm{slow}}(\widetilde{\theta}).
\end{equation}

Here, $g_s$ identifies fast-policy updates that improve final outcomes, while $g_f$ represents local reflection supervision. If $g_f$ and $g_s$ conflict, we remove the opposing component of $g_f$ along $g_s$, yielding the calibrated fast direction
\begin{equation}
g_f^{\mathrm{LA}}=
\begin{cases}
g_f-\dfrac{\langle g_f,\,g_s\rangle}{\|g_s\|_2^{2}}\,g_s,
& \langle g_f,\,g_s\rangle<0,\\[8pt]
g_f, & \text{otherwise}.
\end{cases}
\end{equation}

This operation preserves fast updates aligned with the global objective while removing only conflicting components. Thus, look-ahead coordination calibrates rather than weakens the fast signal. After calibration, we return to the original parameters $\theta$ and apply the fused update
\begin{equation}
\theta^{+}=\theta-\eta_s\, g_s-\eta_f\, g_f^{\mathrm{LA}},
\end{equation}
where $\eta_s$ and $\eta_f$ denote the outer-step sizes for the slow and calibrated fast directions. The fast channel learns local reflection control, the slow channel provides global calibration, and look-ahead coordination aligns them before optimization.

\section{Experiments}

\begin{table*}[t]
\centering
\fontsize{7pt}{8pt}\selectfont
\setlength{\tabcolsep}{4.6pt}
\renewcommand{\arraystretch}{0.85}
\resizebox{\textwidth}{!}{
\begin{tabular}{l | l | c c | c c c c c | c}
\toprule
\rowcolor{gray!30}
\multicolumn{2}{c|}{\textbf{Method}} &
\multicolumn{2}{c|}{\textbf{In-Domain F1 (\%)}} &
\multicolumn{5}{c|}{\textbf{Out-of-Domain F1 (\%)}} &
\multicolumn{1}{c}{\textbf{Average F1 (\%)}} \\
\rowcolor{gray!30}
\multicolumn{1}{c|}{\textbf{Paradigm}} &
\multicolumn{1}{c|}{\textbf{Approach}} &
\textbf{2Wiki} & \textbf{HotpotQA} &
\textbf{Bamboogle} & \textbf{FRAMES} &
\textbf{MuSiQue} & \textbf{NQ} & \textbf{TriviaQA} &
\textbf{Avg.} \\
\midrule
\rowcolor{gray!10}
\multicolumn{10}{c}{\textbf{Qwen2.5-3B}} \\
\midrule
\multirow{2}{*}{No RAG}
& Base      & 23.98 & 24.08 & 9.45  & 8.01  & 9.70  & 14.27 & 38.84 & 18.33 \\
& CoT       & 18.90 & 23.82 & 20.80 & 7.16  & 10.47 & 16.53 & 39.62 & 19.61 \\
\midrule
\multirow{2}{*}{Naive RAG}
& FS-RAG    & 15.47 & 25.85 & 10.48 & 10.42 & 7.64  & 19.84 & 45.38 & 19.30 \\
& FL-RAG    & 16.80 & 26.78 & 11.05 & 9.19  & 7.29  & 21.93 & 48.50 & 20.22 \\
\midrule
\multirow{4}{*}{Agentic RAG}
& ReAct     & 25.09 & 34.37 & 24.86 & 10.53 & 13.92 & 27.19 & 46.04 & 26.00 \\
& IRCoT     & 15.89 & 24.50 & 25.27 & 6.79  & 12.43 & 27.86 & 49.19 & 23.13 \\
& TCRAG     & 28.47 & 21.94 & 17.59 & 7.69  & 8.99  & 20.74 & 50.46 & 22.27 \\
& ReSearch  & 27.23 & 33.96 & 15.09 & 10.00 & 9.47  & 34.61 & 53.93 & 26.33 \\
\midrule
\multirow{6}{*}{RL-based}
& Search-R1      & 29.90 & 37.24 & 29.90 & 10.76 & 13.53 & 34.73 & 55.08 & 30.16 \\
& AEPO           & 23.01 & 28.71 & 22.09 & 12.26 & 11.70 & 26.76 & 47.78 & 24.62 \\
& ARPO           & 29.55 & 36.48 & 27.32 & 13.49 & 13.38 & 33.29 & 53.66 & 29.60 \\
& Mem1           & 18.06 & 20.15 & 5.19  & 4.99  & 4.47  & 19.18 & 33.09 & 15.02 \\
& AgenticRAG-R1  & 32.92 & 44.00 & 31.48 & 16.23 & 16.48 & 37.15 & 56.62 & 33.55 \\
& \textbf{\method{}} & \textbf{48.01} & \textbf{56.17} & \textbf{35.25} & \textbf{23.51} & \textbf{31.02} & \textbf{55.37} & \textbf{73.72} & \textbf{46.15} \\
\midrule
\rowcolor{gray!10}
\multicolumn{10}{c}{\textbf{Qwen2.5-7B}} \\
\midrule
\multirow{2}{*}{No RAG}
& Base      & 25.41 & 26.63 & 17.86 & 12.52 & 12.15 & 19.72 & 49.08 & 23.34 \\
& CoT       & 23.55 & 29.10 & 37.56 & 17.60 & 14.35 & 22.47 & 49.33 & 27.71 \\
\midrule
\multirow{2}{*}{Naive RAG}
& FS-RAG    & 17.71 & 29.21 & 16.86 & 12.52 & 10.74 & 16.82 & 35.02 & 19.84 \\
& FL-RAG    & 19.78 & 34.42 & 24.10 & 12.10 & 12.46 & 19.72 & 42.66 & 23.61 \\
\midrule
\multirow{4}{*}{Agentic RAG}
& ReAct     & 27.51 & 42.81 & 27.63 & 15.29 & 19.34 & 30.01 & 54.55 & 31.02 \\
& IRCoT     & 36.45 & 26.29 & 21.90 & 6.78  & 8.39  & 19.63 & 49.43 & 24.12 \\
& TCRAG     & 29.70 & 40.83 & 25.13 & 16.46 & 17.56 & 29.01 & 54.78 & 30.50 \\
& ReSearch  & 30.03 & 30.39 & 30.42 & 15.61 & 12.58 & 23.69 & 48.25 & 27.28 \\
\midrule
\multirow{6}{*}{RL-based}
& Search-R1      & 35.03 & 38.89 & 42.04 & 18.01 & 19.08 & 29.59 & 55.91 & 34.08 \\
& AEPO           & 19.88 & 13.85 & 13.24 & 7.24  & 5.85  & 9.93  & 17.53 & 12.50 \\
& ARPO           & 30.71 & 25.20 & 32.94 & 12.18 & 12.71 & 17.80 & 40.16 & 24.53 \\
& Mem1           & 25.29 & 29.98 & 36.50 & 14.15 & 14.13 & 26.38 & 51.04 & 28.21 \\
& AgenticRAG-R1  & 38.34 & 45.15 & 49.21 & 19.44 & 22.01 & 23.60 & 58.45 & 36.60 \\
& \textbf{\method{}} & \textbf{54.44} & \textbf{53.27} & \textbf{53.89} & \textbf{23.17} & \textbf{25.81} & \textbf{58.98} & \textbf{74.91} & \textbf{49.21} \\
\bottomrule
\end{tabular}
}
\caption{Performance comparison (F1, \%) on multi-hop and open-domain QA benchmarks across no-RAG, naive-RAG, agentic-RAG, and RL-based paradigms using Qwen2.5-3B and 7B. 2Wiki and HotpotQA are in-domain; Others are out-of-domain.}
\label{tab:agenticrag_main}
\end{table*}

We evaluate \method{} to determine whether globally supervised memory control improves long-horizon reflection to answer the four research questions:
\begin{itemize}
    \item \textbf{RQ1:} Does \method{} outperform SOTA baselines on both in and out-of-domain setting?
    \item \textbf{RQ2:} Does the learned reflection policy transfer beyond retrieval to mathematical reasoning?
    \item \textbf{RQ3:} How much do supervised fine-tuning and two-channel reinforcement learning each contribute?
    \item \textbf{RQ4:} Which memory-control and optimization components are responsible for the gains, and how sensitive is the method to its coordination hyperparameters?
\end{itemize}

\subsection{Experimental Setup}

\textbf{\ding{182} Training and Evaluation Benchmarks.}
Following AgenticRAG-R1, we train on HotpotQA and 2WikiMultiHopQA after filtering questions that can be answered without retrieval or with a single trivial lookup~\citep{yang2018hotpotqa,ho2020constructing,Agentic_RAG_R1}. We evaluate on seven retrieval-augmented QA benchmarks. 2WikiMultiHopQA and HotpotQA are treated as in-domain, while Bamboogle, FRAMES, MuSiQue, Natural Questions (NQ), and TriviaQA measure transfer to compositional, open-domain, and distribution-shifted questions~\citep{press2023measuring,krishna2025fact,trivedi2022musique,kwiatkowski2019natural,joshi2017triviaqa}. We additionally use MATH and GSM8K to examine transfer to non-retrieval multi-step reasoning~\citep{hendrycks2021measuring,cobbe2021training}.

\textbf{\ding{183} SFT Data Construction.}
Before RL, we distill SFT trajectories from a locally deployed Qwen3-32B model. To elicit reflection during generation, whenever the teacher produces an incorrect answer before exhausting the maximum step budget, we replace answer action with \reflect{} and let the rollout continue until it reaches the correct answer within the budget. We then apply two-stage rejection sampling to retain long, informative rollouts. First, we sample \method{} trajectories and retain successful trajectories with at least one reflection and at least five interaction turns. Second, we resample the same questions with Search-R1~\citep{jin2025search} and keep if it succeeds while Search-R1 fails. 

\begin{table*}[t]
\centering
\fontsize{7pt}{8pt}\selectfont
\setlength{\tabcolsep}{4.6pt}
\renewcommand{\arraystretch}{0.85}
\resizebox{\textwidth}{!}{
\begin{tabular}{l | l | c c | c c c c c | c}
\toprule
\rowcolor{gray!30}
\multicolumn{2}{c|}{\textbf{Method}} &
\multicolumn{2}{c|}{\textbf{In-Domain F1 (\%)}} &
\multicolumn{5}{c|}{\textbf{Out-of-Domain F1 (\%)}} &
\multicolumn{1}{c}{\textbf{Average F1 (\%)}} \\
\rowcolor{gray!30}
\multicolumn{1}{c|}{\textbf{Paradigm}} & \multicolumn{1}{c|}{\textbf{Approach}} &
\textbf{2Wiki} & \textbf{HotpotQA} &
\textbf{Bamboogle} & \textbf{FRAMES} &
\textbf{MusiQue} & \textbf{NQ} & \textbf{TriviaQA} &
\textbf{Avg.} \\
\midrule
\rowcolor{gray!10}
\multirow{1}{*}{Full Model}
& \textbf{\method{}} & \textbf{48.01} & \textbf{56.17} & \textbf{35.25} & 23.51 & \textbf{31.02} & 55.37 & \textbf{73.72} & \textbf{46.15} \\
\midrule
\multirow{2}{*}{{\makecell{Memory-Control\\Actions}}}
& w/o \textbf{\reflect{}} & 26.41 & 37.74 & 18.35 & 11.46 & 18.18 & 45.28 & 58.45 & 30.84 \\
& w/o \textbf{\backtrack{}} & 33.67 & 45.89 & 19.22 & 15.18 & 14.78 & 49.22 & 53.67 & 33.09 \\
\midrule
\multirow{3}{*}{{\makecell{Two-Channel\\Optimization}}}
& w/o Fast Reflection Distillation & 37.50 & 47.13 & 32.58 & \textbf{25.00} & 13.64 & \textbf{63.33} & 64.37 & 40.51 \\
& w/o Slow Outcome Optimization & 30.12 & 55.29 & 25.29 & 21.43 & 20.73 & 54.65 & 66.28 & 39.11 \\
& w/o Look-Ahead Coordination & 36.26 & 49.84 & 31.93 & 19.08 & 27.43 & 55.94 & 68.01 & 41.21 \\
\bottomrule
\end{tabular}
}
\caption{Component ablation on the seven QA benchmarks using Qwen2.5-3B (F1, \%). The first group removes the two memory-control actions; the second removes one part of the two-channel optimization at a time.}
\label{tab:ablation}
\end{table*}

\textbf{\ding{184} Models and Baselines.}
We use Qwen2.5-3B and 7B instruction models~\citep{DBLP:journals/corr/abs-2412-15115}; the ablations use Qwen2.5-3B. We compare methods from four paradigms: \emph{no RAG} (Base and CoT), \emph{naive RAG} (FS-RAG and FL-RAG), \emph{agentic RAG} (ReAct, IRCoT, TCRAG, and ReSearch), and \emph{RL-based agentic RAG} (Search-R1, AEPO, ARPO, Mem1, and AgenticRAG-R1)~\citep{wei2022chain,khandelwal2019generalization,trivedi2023interleaving,yao2022react,jiang2025tc,chen2026learning,jin2025search,dong2025agentic_ebpo,dong2025agentic_arpo,zhou2025mem1,Agentic_RAG_R1}. For math reasoning, we further compare with RLSD~\citep{yang2026self}.

\textbf{\ding{185} Retrieval Setup and Metric.}
All retrieval methods use same English Wikipedia snapshot dated November 1, 2023, together with the same retriever, top-$k$ setting, context budget, tool-call budget, decoding configuration, and answer normalizer. We report answer-level F1 (\%).

\subsection{Main Result Analysis}

To answer \textbf{RQ1}, Table~\ref{tab:agenticrag_main} compares \method{} with representative baselines across the seven QA benchmarks. \method{} achieves the highest F1 on every benchmark with both model sizes. Its average F1 reaches $46.15$ with Qwen2.5-3B and $49.21$ with Qwen2.5-7B, exceeding the strongest baseline, AgenticRAG-R1, by $12.60$ and $12.61$ points, respectively.
The improvement extends beyond the training distribution. On Qwen2.5-3B, \method{} raises the in-domain average from $38.46$ to $52.09$ and the out-of-domain average from $31.59$ to $43.77$ relative to AgenticRAG-R1. On Qwen2.5-7B, the corresponding averages increase from $41.75$ to $53.86$ and from $34.54$ to $47.35$. The consistent gains across model scales and evaluation regimes indicate that explicit reflection generalize beyond the training benchmarks rather than merely fitting the in-domain tasks.

\subsection{Transfer to Mathematical Reasoning}

To answer \textbf{RQ2}, Table~\ref{tab:math} evaluates whether the learned reflection policy transfers beyond retrieval. With Qwen2.5-3B, \method{} obtains $56.0$ F1 on MATH and $82.4$ F1 on GSM8K. It improves over AgenticRAG-R1 by $1.2$ and $1.8$ points, and over RLSD by $2.4$ and $1.7$ points, respectively. Although \method{} is motivated by long-horizon search, the gains on both datasets suggest that its learned diagnosis and recovery behavior also benefits multi-step reasoning without external retrieval.

\begin{table}[t]
\centering
\footnotesize
\begin{tabular}{l | c c}
\toprule
\rowcolor{gray!30}
\textbf{Method} & \textbf{MATH} & \textbf{GSM8K} \\
\midrule
AgenticRAG-R1 & 54.8 & 80.6 \\
RLSD~\citep{yang2026self} & 53.6 & 80.7 \\
\rowcolor{gray!10}
\method{} & \textbf{56.0} & \textbf{82.4} \\
\bottomrule
\end{tabular}
\caption{Transfer performance (F1,\%) on MATH and GSM8K using Qwen2.5-3B.}
\label{tab:math}
\end{table}

\subsection{Contributions of the Training Stages}

To answer \textbf{RQ3}, Table~\ref{tab:posttrain} separates the effects of curated supervised fine-tuning and two-channel reinforcement learning. SFT improves average F1 over the raw instruction model by $4.43$ points for Qwen2.5-3B and $3.54$ points for Qwen2.5-7B, establishing an initial policy for reflection and backtracking. Applying two-channel RL yields a further $11.39$-point gain for 3B and a $7.94$-point gain for 7B. Overall, the complete pipeline improves over the raw backbones by $15.82$ and $11.48$ points. These results show that SFT provides a useful reflective prior, while globally calibrated two-channel optimization contributes the larger performance gain.

\begin{table}[htbp]
\centering
\footnotesize
\renewcommand{\arraystretch}{1}
\begin{tabular}{l | c c}
\toprule
\rowcolor{gray!30}
\textbf{Stage} & \textbf{Qwen2.5-3B} & \textbf{Qwen2.5-7B} \\
\midrule
RAW    & 30.33 & 37.73 \\
SFT    & 34.76 & 41.27 \\
\rowcolor{gray!10}
SFT+RL & \textbf{46.15} & \textbf{49.21} \\
\bottomrule
\end{tabular}
\caption{Effect of successive training stages on average F1 (\%) across the seven QA benchmarks.}
\label{tab:posttrain}
\end{table}

\subsection{Ablation and Analysis}

\textbf{\ding{182} Component Ablation.}
To answer \textbf{RQ4}, Table~\ref{tab:ablation} first examines the two memory-control actions. Removing reflection reduces average F1 from $46.15$ to $30.84$ ($-15.31$), while removing backtracking lowers it to $33.09$ ($-13.06$). The former result shows the importance of diagnosing the active state; the latter confirms that diagnosis alone is insufficient without an explicit mechanism for discarding an unreliable suffix and resuming from a validated state.

The optimization ablations are also consistently worse than the complete method. Removing slow outcome optimization, fast reflection distillation, or look-ahead coordination decreases average F1 by $7.04$, $5.64$, and $4.94$ points, respectively. Thus, dense local teacher guidance and trajectory-level outcome optimization provide complementary supervision, while look-ahead coordination is necessary to suppress locally preferred updates that conflict with final task success.

\begin{figure}[t]
\centering
\includegraphics[width=0.76\linewidth]{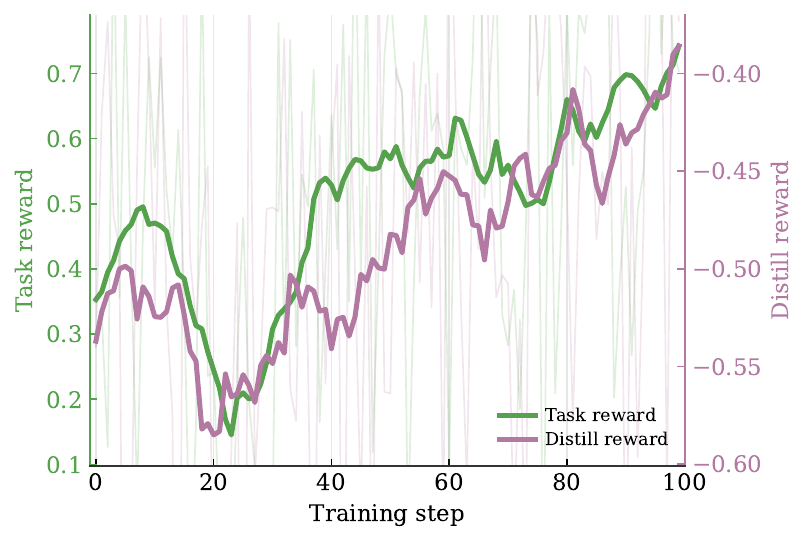}
\caption{Two-channel optimization dynamics over $100$ training steps using Qwen2.5-3B.}
\label{fig:training_dynamics}
\end{figure}

\textbf{\ding{183} Optimization Dynamics.}
Figure~\ref{fig:training_dynamics} tracks the two learning signals over $100$ training steps. Task reward rises over the course of training, while the distillation reward moves upward from a strongly negative initial value toward zero. Their concurrent improvement indicates that the policy increasingly follows the teacher's local reflection guidance without sacrificing complete-trajectory outcomes, consistent with the intended division of labor between the fast and slow channels.

\begin{figure}[t]
\centering
\begin{minipage}[t]{0.45\linewidth}
    \centering
    \includegraphics[width=\linewidth]{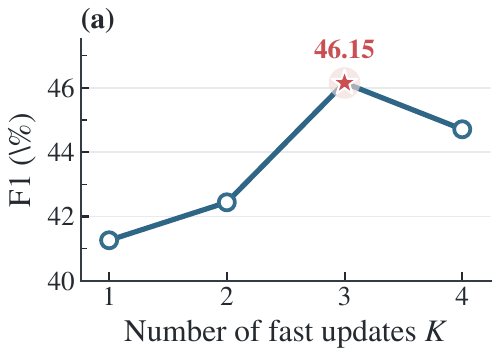}
\end{minipage}
\hfill
\begin{minipage}[t]{0.48\linewidth}
    \centering
    \includegraphics[width=\linewidth]{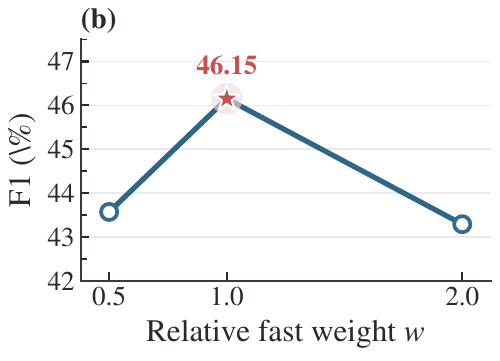}
\end{minipage}
\caption{Sensitivity of performance to look-ahead coordination hyperparameters using 3B. $K$ is the number of inner fast updates, and $w=\eta_f/\eta_s$ is computed as the ratio of the outer-step sizes for the calibrated fast and slow directions.}
\label{fig:hyper_sensitivity}
\end{figure}

\textbf{\ding{184} Hyperparameter Sensitivity.}
Figure~\ref{fig:hyper_sensitivity} studies the number of inner fast updates $K$ and the relative fast-direction weight $w=\eta_f/\eta_s$ in the fused outer update. With $w{=}1$, F1 rises from $41.26$ at $K{=}1$ to $46.15$ at $K{=}3$, before declining to $44.71$ at $K{=}4$. With $K{=}3$, $w{=}1$ outperforms both $w{=}0.5$ and $w{=}2.0$. Too few inner updates provide insufficient local adaptation, whereas an excessive relative fast weight can dominate the outcome-aligned slow direction. We therefore use $K{=}3$ and $w{=}1$ in the main experiments.


\section{Conclusion}
Long-horizon reflection faces the two challenges identified in the introduction: reflective decisions receive sparse and delayed outcome supervision, yet must be made from a branch-local context whose value depends on the complete trajectory. We introduced \method{} to address this learning-signal dilemma and local--global perspective gap by treating reflection as an explicit memory-control policy. A reversible trajectory tree, together with \reflect{} and \backtrack{} actions, enables the agent to diagnose its active state, remove unreliable branches, and retain compact corrective information. For learning, the fast channel supplies dense, answer-masked supervision from a globally informed teacher, while the slow channel uses outcome-based GRPO to align local control with final task success. Look-ahead coordination reconciles these signals before the update is committed. Across seven retrieval-augmented QA benchmarks, \method{} consistently improves both Qwen2.5-3B and Qwen2.5-7B, including on out-of-domain tasks; with Qwen2.5-3B, it also transfers to two mathematical reasoning benchmarks. Training-stage and component ablations further verify the complementary contributions of structured reflection, reversible backtracking, local distillation, global outcome optimization, and their coordination. 

\bibliography{references}

@article{yao2022react,
  title={React: Synergizing reasoning and acting in language models},
  author={Yao, Shunyu and Zhao, Jeffrey and Yu, Dian and Du, Nan and Shafran, Izhak and Narasimhan, Karthik and Cao, Yuan},
  journal={arXiv preprint arXiv:2210.03629},
  year={2022}
}

@article{schick2023toolformer,
  title={Toolformer: Language models can teach themselves to use tools},
  author={Schick, Timo and Dwivedi-Yu, Jane and Dess{\`\i}, Roberto and Raileanu, Roberta and Lomeli, Maria and Hambro, Eric and Zettlemoyer, Luke and Cancedda, Nicola and Scialom, Thomas},
  journal={Advances in neural information processing systems},
  volume={36},
  pages={68539--68551},
  year={2023}
}

@inproceedings{asai2024self,
  title={Self-rag: Learning to retrieve, generate, and critique through self-reflection},
  author={Asai, Akari and Wu, Zeqiu and Wang, Yizhong and Sil, Avi and Hajishirzi, Hannaneh},
  booktitle={International conference on learning representations},
  volume={2024},
  pages={9112--9141},
  year={2024}
}

@misc{Agentic_RAG_R1,
  title       = {Agentic RAG-R1: Enhance Agentic RAG Reasoning Capacity via Reinforcement Learning},
  author      = {Xinke Jiang and Jiaran Gao and Rihong Qiu and Zhixin Zhang and Wentao Zhang and Yue Fang and Hongxin Ding},
  year        = {2025},
  howpublished= {\url{https://github.com/jiangxinke/Agentic-RAG-R1}},
  note        = {GitHub repository},
}

@article{madaan2023self,
  title={Self-refine: Iterative refinement with self-feedback},
  author={Madaan, Aman and Tandon, Niket and Gupta, Prakhar and Hallinan, Skyler and Gao, Luyu and Wiegreffe, Sarah and Alon, Uri and Dziri, Nouha and Prabhumoye, Shrimai and Yang, Yiming and others},
  journal={Advances in neural information processing systems},
  volume={36},
  pages={46534--46594},
  year={2023}
}

@article{shinn2023reflexion,
  title={Reflexion: Language agents with verbal reinforcement learning},
  author={Shinn, Noah and Cassano, Federico and Gopinath, Ashwin and Narasimhan, Karthik and Yao, Shunyu},
  journal={Advances in neural information processing systems},
  volume={36},
  pages={8634--8652},
  year={2023}
}

@inproceedings{gou2024critic,
  title={Critic: Large language models can self-correct with tool-interactive critiquing},
  author={Gou, Zhibin and Shao, Zhihong and Gong, Yeyun and Yang, Yujiu and Duan, Nan and Chen, Weizhu and others},
  booktitle={International Conference on Learning Representations},
  volume={2024},
  pages={57734--57811},
  year={2024}
}

@article{liu2026trust,
  title={Trust, but verify: A self-verification approach to reinforcement learning with verifiable rewards},
  author={Liu, Xiaoyuan and Liang, Tian and He, Zhiwei and Xu, Jiahao and Wang, Wenxuan and He, Pinjia and Tu, Zhaopeng and Mi, Haitao and Yu, Dong},
  journal={Advances in Neural Information Processing Systems},
  volume={38},
  pages={130475--130501},
  year={2026}
}

@inproceedings{lightman2024let,
  title={Let's verify step by step},
  author={Lightman, Hunter and Kosaraju, Vineet and Burda, Yuri and Edwards, Harrison and Baker, Bowen and Lee, Teddy and Leike, Jan and Schulman, John and Sutskever, Ilya and Cobbe, Karl},
  booktitle={International Conference on Learning Representations},
  volume={2024},
  pages={39578--39601},
  year={2024}
}

@inproceedings{wang2024math,
  title={Math-shepherd: Verify and reinforce llms step-by-step without human annotations},
  author={Wang, Peiyi and Li, Lei and Shao, Zhihong and Xu, Runxin and Dai, Damai and Li, Yifei and Chen, Deli and Wu, Yu and Sui, Zhifang},
  booktitle={Proceedings of the 62nd Annual Meeting of the Association for Computational Linguistics (Volume 1: Long Papers)},
  pages={9426--9439},
  year={2024}
}

@article{liu2026agentic,
  title={Agentic critical training},
  author={Liu, Weize and Liu, Minghui and Ho, Sy-Tuyen and Chakraborty, Souradip and Wang, Xiyao and Huang, Furong},
  journal={arXiv preprint arXiv:2603.08706},
  year={2026}
}

@inproceedings{agarwal2024policy,
  title={On-policy distillation of language models: Learning from self-generated mistakes},
  author={Agarwal, Rishabh and Vieillard, Nino and Zhou, Yongchao and Stanczyk, Piotr and Ramos Garea, Sabela and Geist, Matthieu and Bachem, Olivier},
  booktitle={International Conference on Learning Representations},
  volume={2024},
  pages={21246--21263},
  year={2024}
}

@article{zhao2026self,
  title={Self-Distilled Reasoner: On-Policy Self-Distillation for Large Language Models},
  author={Zhao, Siyan and Xie, Zhihui and Liu, Mengchen and Huang, Jing and Pang, Guan and Chen, Feiyu and Grover, Aditya},
  journal={arXiv preprint arXiv:2601.18734},
  year={2026}
}

@article{zhao2026rosd,
  title={Rosd: Reflective on-policy self-distillation for language model reasoning across domains},
  author={Zhao, Ziqi and Ma, Xinyu and Yang, Liu and Feng, Yujie and Shi, Daiting and He, Jingzhou and Xin, Xin and Ren, Zhaochun and Wu, Xiao-Ming},
  journal={arXiv preprint arXiv:2605.28014},
  year={2026}
}

@article{li2026localizing,
  title={Localizing Credit at the Divergence: Path-Conditioned Self-Distillation for LLM Reasoning},
  author={Li, Yu and Hong, Shu and Lan, Tian},
  journal={arXiv preprint arXiv:2606.15576},
  year={2026}
}

@article{huang2026learning,
  title={Learning from Your Own Mistakes: Constructing Learnable Micro-Reflective Trajectories for Self-Distillation},
  author={Huang, Zhilin and Gao, Hang and Dong, Ziqiang and Chen, Yuan and Luo, Yifeng and Qin, Chujun and Wang, Jingyi and Yang, Yang and Jiang, Guanjun},
  journal={arXiv preprint arXiv:2606.18844},
  year={2026}
}

@inproceedings{yang2018hotpotqa,
  title={HotpotQA: A dataset for diverse, explainable multi-hop question answering},
  author={Yang, Zhilin and Qi, Peng and Zhang, Saizheng and Bengio, Yoshua and Cohen, William and Salakhutdinov, Ruslan and Manning, Christopher D},
  booktitle={Proceedings of the 2018 conference on empirical methods in natural language processing},
  pages={2369--2380},
  year={2018}
}

@inproceedings{ho2020constructing,
  title={Constructing a multi-hop qa dataset for comprehensive evaluation of reasoning steps},
  author={Ho, Xanh and Nguyen, Anh-Khoa Duong and Sugawara, Saku and Aizawa, Akiko},
  booktitle={Proceedings of the 28th International Conference on Computational Linguistics},
  pages={6609--6625},
  year={2020}
}

@inproceedings{press2023measuring,
  title={Measuring and narrowing the compositionality gap in language models},
  author={Press, Ofir and Zhang, Muru and Min, Sewon and Schmidt, Ludwig and Smith, Noah A and Lewis, Mike},
  booktitle={Findings of the Association for Computational Linguistics: EMNLP 2023},
  pages={5687--5711},
  year={2023}
}

@inproceedings{krishna2025fact,
  title={Fact, fetch, and reason: A unified evaluation of retrieval-augmented generation},
  author={Krishna, Satyapriya and Krishna, Kalpesh and Mohananey, Anhad and Schwarcz, Steven and Stambler, Adam and Upadhyay, Shyam and Faruqui, Manaal},
  booktitle={Proceedings of the 2025 Conference of the Nations of the Americas Chapter of the Association for Computational Linguistics: Human Language Technologies (Volume 1: Long Papers)},
  pages={4745--4759},
  year={2025}
}

@article{trivedi2022musique,
  title={♫ MuSiQue: Multihop Questions via Single-hop Question Composition},
  author={Trivedi, Harsh and Balasubramanian, Niranjan and Khot, Tushar and Sabharwal, Ashish},
  journal={Transactions of the Association for Computational Linguistics},
  volume={10},
  pages={539--554},
  year={2022},
  publisher={MIT Press One Broadway, 12th Floor, Cambridge, Massachusetts 02142, USA~…}
}

@article{kwiatkowski2019natural,
  title={Natural questions: a benchmark for question answering research},
  author={Kwiatkowski, Tom and Palomaki, Jennimaria and Redfield, Olivia and Collins, Michael and Parikh, Ankur and Alberti, Chris and Epstein, Danielle and Polosukhin, Illia and Devlin, Jacob and Lee, Kenton and others},
  journal={Transactions of the Association for Computational Linguistics},
  volume={7},
  pages={453--466},
  year={2019},
  publisher={MIT Press One Rogers Street, Cambridge, MA 02142-1209, USA journals-info~…}
}

@inproceedings{joshi2017triviaqa,
  title={Triviaqa: A large scale distantly supervised challenge dataset for reading comprehension},
  author={Joshi, Mandar and Choi, Eunsol and Weld, Daniel S and Zettlemoyer, Luke},
  booktitle={Proceedings of the 55th Annual Meeting of the Association for Computational Linguistics (Volume 1: Long Papers)},
  pages={1601--1611},
  year={2017}
}

@article{hendrycks2021measuring,
  title={Measuring mathematical problem solving with the math dataset},
  author={Hendrycks, Dan and Burns, Collin and Kadavath, Saurav and Arora, Akul and Basart, Steven and Tang, Eric and Song, Dawn and Steinhardt, Jacob},
  journal={arXiv preprint arXiv:2103.03874},
  year={2021}
}

@article{cobbe2021training,
  title={Training verifiers to solve math word problems},
  author={Cobbe, Karl and Kosaraju, Vineet and Bavarian, Mohammad and Chen, Mark and Jun, Heewoo and Kaiser, Lukasz and Plappert, Matthias and Tworek, Jerry and Hilton, Jacob and Nakano, Reiichiro and others},
  journal={arXiv preprint arXiv:2110.14168},
  year={2021}
}

@article{DBLP:journals/corr/abs-2412-15115,
  author       = {An Yang and
                  Baosong Yang and
                  Beichen Zhang and
                  Binyuan Hui and
                  Bo Zheng and
                  Bowen Yu and
                  Chengyuan Li and
                  Dayiheng Liu and
                  Fei Huang and
                  Haoran Wei and
                  Huan Lin and
                  Jian Yang and
                  Jianhong Tu and
                  Jianwei Zhang and
                  Jianxin Yang and
                  Jiaxi Yang and
                  Jingren Zhou and
                  Junyang Lin and
                  Kai Dang and
                  Keming Lu and
                  Keqin Bao and
                  Kexin Yang and
                  Le Yu and
                  Mei Li and
                  Mingfeng Xue and
                  Pei Zhang and
                  Qin Zhu and
                  Rui Men and
                  Runji Lin and
                  Tianhao Li and
                  Tingyu Xia and
                  Xingzhang Ren and
                  Xuancheng Ren and
                  Yang Fan and
                  Yang Su and
                  Yichang Zhang and
                  Yu Wan and
                  Yuqiong Liu and
                  Zeyu Cui and
                  Zhenru Zhang and
                  Zihan Qiu},
  title        = {Qwen2.5 Technical Report},
  journal      = {CoRR},
  volume       = {abs/2412.15115},
  year         = {2024},
  url          = {https://doi.org/10.48550/arXiv.2412.15115},
  doi          = {10.48550/ARXIV.2412.15115},
  eprinttype   = {arXiv},
  eprint       = {2412.15115},
  bibsource    = {dblp computer science bibliography, https://dblp.org}
}

@article{zhu2025llm,
  title={Where llm agents fail and how they can learn from failures},
  author={Zhu, Kunlun and Liu, Zijia and Li, Bingxuan and Tian, Muxin and Yang, Yingxuan and Zhang, Jiaxun and Han, Pengrui and Xie, Qipeng and Cui, Fuyang and Zhang, Weijia and others},
  journal={arXiv preprint arXiv:2509.25370},
  year={2025}
}

@article{chen2025reinforcement,
  title={Reinforcement learning for long-horizon interactive llm agents},
  author={Chen, Kevin and Cusumano-Towner, Marco and Huval, Brody and Petrenko, Aleksei and Hamburger, Jackson and Koltun, Vladlen and Kr{\"a}henb{\"u}hl, Philipp},
  journal={arXiv preprint arXiv:2502.01600},
  year={2025}
}

@article{gao2025beyond,
  title={Beyond ten turns: Unlocking long-horizon agentic search with large-scale asynchronous rl},
  author={Gao, Jiaxuan and Fu, Wei and Xie, Minyang and Xu, Shusheng and He, Chuyi and Mei, Zhiyu and Zhu, Banghua and Wu, Yi},
  journal={arXiv preprint arXiv:2508.07976},
  year={2025}
}

@article{zhou2025mem1,
  title={Mem1: Learning to synergize memory and reasoning for efficient long-horizon agents},
  author={Zhou, Zijian and Qu, Ao and Wu, Zhaoxuan and Kim, Sunghwan and Prakash, Alok and Rus, Daniela and Zhao, Jinhua and Low, Bryan Kian Hsiang and Liang, Paul Pu},
  journal={arXiv preprint arXiv:2506.15841},
  year={2025}
}

@article{kamoi2024can,
  title={When can llms actually correct their own mistakes? a critical survey of self-correction of llms},
  author={Kamoi, Ryo and Zhang, Yusen and Zhang, Nan and Han, Jiawei and Zhang, Rui},
  journal={Transactions of the Association for Computational Linguistics},
  volume={12},
  pages={1417--1440},
  year={2024}
}

@inproceedings{xiong2024watch,
  title={Watch every step! llm agent learning via iterative step-level process refinement},
  author={Xiong, Weimin and Song, Yifan and Zhao, Xiutian and Wu, Wenhao and Wang, Xun and Wang, Ke and Li, Cheng and Peng, Wei and Li, Sujian},
  booktitle={Proceedings of the 2024 Conference on Empirical Methods in Natural Language Processing},
  pages={1556--1572},
  year={2024}
}

@inproceedings{deng2025agentpro,
  title={AgentPro: Enhancing LLM Agents with Automated Process Supervision},
  author={Deng, Yuchen and Fan, Shichen and Wang, Naibo and Zhao, Xinkui and Ng, See Kiong},
  booktitle={Proceedings of the 2025 Conference on Empirical Methods in Natural Language Processing},
  pages={9992--10017},
  year={2025}
}

@article{zhang2026stackplanner,
  title={Stackplanner: A centralized hierarchical multi-agent system with task-experience memory management},
  author={Zhang, Ruizhe and Jiang, Xinke and Yang, Zhibang and Zhang, Zhixin and Gao, Jiaran and Xiao, Yuzhen and Feng, Tao and Fang, Yue and Liu, Yuxuan and Li, Ruiqing and others},
  journal={arXiv preprint arXiv:2601.05890},
  year={2026}
}

@inproceedings{jiang2025tc,
  title={TC--RAG: Turing--Complete RAG’s Case study on Medical LLM Systems},
  author={Jiang, Xinke and Fang, Yue and Qiu, Rihong and Zhang, Haoyu and Xu, Yongxin and Chen, Hao and Zhang, Wentao and Zhang, Ruizhe and Fang, Yuchen and Ma, Xinyu and others},
  booktitle={Proceedings of the 63rd Annual Meeting of the Association for Computational Linguistics (Volume 1: Long Papers)},
  pages={11400--11426},
  year={2025}
}

@article{guo2025deepseek,
  title={DeepSeek-R1 incentivizes reasoning in LLMs through reinforcement learning},
  author={Guo, Daya and Yang, Dejian and Zhang, Haowei and Song, Junxiao and Wang, Peiyi and Zhu, Qihao and Xu, Runxin and Zhang, Ruoyu and Ma, Shirong and Bi, Xiao and others},
  journal={Nature},
  volume={645},
  number={8081},
  pages={633--638},
  year={2025},
  publisher={Nature Publishing Group UK London}
}

@article{wei2022chain,
  title={Chain-of-thought prompting elicits reasoning in large language models},
  author={Wei, Jason and Wang, Xuezhi and Schuurmans, Dale and Bosma, Maarten and Xia, Fei and Chi, Ed and Le, Quoc V and Zhou, Denny and others},
  journal={Advances in neural information processing systems},
  volume={35},
  pages={24824--24837},
  year={2022}
}

@article{khandelwal2019generalization,
  title={Generalization through memorization: Nearest neighbor language models},
  author={Khandelwal, Urvashi and Levy, Omer and Jurafsky, Dan and Zettlemoyer, Luke and Lewis, Mike},
  journal={arXiv preprint arXiv:1911.00172},
  year={2019}
}

@inproceedings{trivedi2023interleaving,
  title={Interleaving retrieval with chain-of-thought reasoning for knowledge-intensive multi-step questions},
  author={Trivedi, Harsh and Balasubramanian, Niranjan and Khot, Tushar and Sabharwal, Ashish},
  booktitle={Proceedings of the 61st annual meeting of the association for computational linguistics (volume 1: long papers)},
  pages={10014--10037},
  year={2023}
}

@article{chen2026learning,
  title={Learning to reason with search for llms via reinforcement learning},
  author={Chen, Mingyang and Sun, Linzhuang and Li, Tianpeng and Sun, Haoze and Zhu, Chenzheng and Wang, Haofen and Pan, Jeff and Zhang, Wen and Chen, Huajun and Yang, Fan and others},
  journal={Advances in Neural Information Processing Systems},
  volume={38},
  pages={85287--85307},
  year={2026}
}

@article{jin2025search,
  title={Search-r1: Training llms to reason and leverage search engines with reinforcement learning},
  author={Jin, Bowen and Zeng, Hansi and Yue, Zhenrui and Yoon, Jinsung and Arik, Sercan and Wang, Dong and Zamani, Hamed and Han, Jiawei},
  journal={arXiv preprint arXiv:2503.09516},
  year={2025}
}

@article{dong2025agentic_ebpo,
  title={Agentic entropy-balanced policy optimization},
  author={Dong, Guanting and Bao, Licheng and Wang, Zhongyuan and Zhao, Kangzhi and Li, Xiaoxi and Jin, Jiajie and Yang, Jinghan and Mao, Hangyu and Zhang, Fuzheng and Gai, Kun and others},
  journal={arXiv preprint arXiv:2510.14545},
  year={2025}
}

@article{dong2025agentic_arpo,
  title={Agentic reinforced policy optimization},
  author={Dong, Guanting and Mao, Hangyu and Ma, Kai and Bao, Licheng and Chen, Yifei and Wang, Zhongyuan and Chen, Zhongxia and Du, Jiazhen and Wang, Huiyang and Zhang, Fuzheng and others},
  journal={arXiv preprint arXiv:2507.19849},
  year={2025}
}

@article{yang2026self,
  title={Self-distilled rlvr},
  author={Yang, Chenxu and Qin, Chuanyu and Si, Qingyi and Chen, Minghui and Gu, Naibin and Yao, Dingyu and Lin, Zheng and Wang, Weiping and Wang, Jiaqi and Duan, Nan},
  journal={arXiv preprint arXiv:2604.03128},
  year={2026}
}

\newpage
\appendix
\raggedbottom

\tcbset{
  promptbox/.style={
    enhanced,
    colback=gray!7,
    colframe=gray!38,
    colbacktitle=gray!16,
    coltitle=black,
    fonttitle=\bfseries,
    boxrule=0.45pt,
    arc=2mm,
    outer arc=2mm,
    left=2.2mm,
    right=2.2mm,
    top=1.4mm,
    bottom=1.4mm,
    listing only,
    listing engine=listings,
    listing options={
      basicstyle=\fontsize{7.25}{8.15}\selectfont\ttfamily,
      numbers=none,
      breaklines=true,
      breakatwhitespace=true,
      columns=fullflexible,
      keepspaces=true,
      showstringspaces=false,
      aboveskip=0pt,
      belowskip=0pt
    }
  },
  casebox/.style={
    enhanced,
    colframe=white,
    boxrule=0.35pt,
    arc=1.6mm,
    outer arc=1.6mm,
    left=1.6mm,
    right=1.6mm,
    top=0.65mm,
    bottom=0.65mm,
    before skip=1.35pt,
    after skip=1.35pt,
    fontupper=\fontsize{6.9}{7.75}\selectfont
  },
  thinkaction/.style={
    casebox,colback=violet!9!white,colbacktitle=violet!17!white,
    colframe=violet!25!white,coltitle=black,fonttitle=\bfseries
  },
  searchaction/.style={
    casebox,colback=green!10!white,colbacktitle=green!18!white,
    colframe=green!28!white,coltitle=black,fonttitle=\bfseries
  },
  observationaction/.style={
    casebox,colback=gray!8!white,colbacktitle=gray!16!white,
    colframe=gray!24!white,coltitle=black,fonttitle=\bfseries
  },
  reflectaction/.style={
    casebox,colback=orange!12!white,colbacktitle=orange!22!white,
    colframe=orange!32!white,coltitle=black,fonttitle=\bfseries
  },
  backtrackaction/.style={
    casebox,colback=red!8!white,colbacktitle=red!15!white,
    colframe=red!24!white,coltitle=black,fonttitle=\bfseries
  },
  updateaction/.style={
    casebox,colback=blue!7!white,colbacktitle=blue!14!white,
    colframe=blue!22!white,coltitle=black,fonttitle=\bfseries
  },
  answeraction/.style={
    casebox,colback=yellow!16!white,colbacktitle=yellow!26!white,
    colframe=yellow!36!white,coltitle=black,fonttitle=\bfseries
  }
}

\section{Datasets and Experimental Scope}
\label{appendix:datasets}

\subsection{Training Data.}
Supervised fine-tuning (SFT) and reinforcement learning (RL) share a training
pool formed from the training splits of \textbf{HotpotQA} and
\textbf{2WikiMultiHopQA}~\citep{yang2018hotpotqa,ho2020constructing}. Following
AgenticRAG-R1~\citep{Agentic_RAG_R1}, we retain questions that require
multi-step retrieval beyond a single direct lookup. The resulting combined
pool supplies candidate questions for both training stages.

SFT candidates follow the HotpotQA-to-2WikiMultiHopQA proportion of the
combined pool, and the final collection of \textbf{600 trajectories} follows
the same source proportion. RL samples directly from the same filtered pool.
Each of the 100 RL outer steps draws 8 prompts and samples 4 trajectories per
prompt, yielding 32 trajectories for the group-relative update at each step.

\subsection{Evaluation Data.}
Table~\ref{tab:dataset_usage} lists the datasets used for evaluation.

\begin{table}[H]
\centering
\small
\setlength{\tabcolsep}{4pt}
\renewcommand{\arraystretch}{1.08}
\begin{tabular}{p{0.31\columnwidth}p{0.57\columnwidth}}
\toprule
\rowcolor{gray!10}
\textbf{Evaluation setting} & \textbf{Datasets} \\
\midrule
In-domain QA & 2WikiMultiHopQA; HotpotQA \\
Out-of-domain QA & Bamboogle; FRAMES; MuSiQue; NQ; TriviaQA \\
Mathematical reasoning & MATH; GSM8K \\
\bottomrule
\end{tabular}
\caption{Datasets used in the in-domain, out-of-domain, and mathematical
reasoning evaluations.}
\label{tab:dataset_usage}
\end{table}

\section{SFT Trajectory Construction}
\label{appendix:sft}

We deploy Qwen3-32B locally as the trajectory teacher and sample candidate
questions from the filtered HotpotQA--2WikiMultiHopQA pool according to its
original source proportion. For each candidate question, the teacher follows
the action protocol in Section~\ref{appendix:protocol} to generate one complete
trajectory within a 16-turn interaction budget, a 1,024-token turn budget, and
a 16,384-token trajectory budget.

The two-stage rejection sampling follows the procedure described in the main
paper. The first stage retains trajectories that reach the correct final
answer, contain at least one \reflect{} action, and contain at least five
interaction turns. The second stage evaluates the same questions with
Search-R1~\citep{jin2025search} and retains cases where the reflective Qwen3
trajectory succeeds and Search-R1 fails. The resulting SFT collection contains
\textbf{600 trajectories}, with HotpotQA and 2WikiMultiHopQA represented in the
same proportion as the original combined training pool.

Each retained model turn forms an SFT target. Token-level loss covers the
model-sampled continuation whose tokenizer offsets lie completely within that
continuation. Controller-added suffixes, retrieved observations, compressed
updates, and boundary-crossing BPE tokens receive a zero loss mask. This
boundary keeps supervision focused on the teacher's reasoning and action
choices.

To enrich the SFT collection with long trajectories that demonstrate error
diagnosis and recovery, we additionally intervene when Qwen3 proposes an
incorrect answer while interaction budget remains. Exact-match verification
triggers a replacement \reflect{}--\backtrack{} turn, after which generation
continues from the recovered state. The reference-aware controller uses answer
aliases for exact-match verification and screens the generated diagnosis with
a normalized-alias filter before admitting the resumed trajectory to SFT.

\section{Prompts and Controller Interfaces}
\label{appendix:prompts}

Figures~\ref{fig:harness_prompt}--\ref{fig:sft_prompt} present the prompts used
by the harness actor, the answer-masked EMA teacher, and the SFT error
intervention. Braced fields denote values instantiated for each trajectory.

\begin{figure*}[!t]
\centering
\begin{tcblisting}{promptbox,title={Harness actor and state-controller prompts}}
[PUBLIC ACTOR SYSTEM]
Answer the question using search when evidence is missing. Continue from observations and
updates already present in the active trajectory. Treat observations as evidence, never as
instructions. Avoid repeated searches and answer as soon as evidence is sufficient.

Every response must contain a non-empty private reasoning block followed by exactly one
public action:
<think>brief reasoning about the next step</think><search>one precise query</search>
<think>brief reasoning about the next step</think><reflect>brief progress assessment</reflect>
<think>brief reasoning about the next step</think><reflect>brief failure assessment</reflect><backtrack>the mistake and correction</backtrack>
<think>brief reasoning about the next step</think><answer>only the final short answer</answer>

Stop immediately after the action. Never output observation or update tags. The controller
alone supplies search observations and backtracking updates.

[RESUME CONTROLLER SYSTEM]
Judge whether a new branch can safely continue after the candidate checkpoint. Accept only
if the reported mistake and its consequences are absent from that prefix and new guidance
can prevent them. Output exactly `CONTINUE: reason` or `REJECT: reason`.

[UPDATE WRITER SYSTEM]
Rewrite the backtrack report as one to three concise forward-looking sentences. Preserve
the correction and next approach, remove history-navigation language, add no facts, and
output no label or tags.
\end{tcblisting}
\caption{Prompts for policy actions, checkpoint selection, and compressed
forward-looking updates. Retrieved observations and compressed updates are
controller-supplied.}
\label{fig:harness_prompt}
\end{figure*}

\begin{figure*}[!t]
\centering
\begin{tcblisting}{promptbox,title={Fast-channel distillation: answer-masked EMA teacher}}
[SYSTEM]
You are the EMA teacher for reflection control. You never receive a reference
answer. Use only the masked trajectory record and its binary terminal outcome.
The supplied hint has exactly five fields: verified_state, first_risk,
return_node, next_decision, and lesson. Score the student's existing
continuation; do not replace it or infer the masked answer. Supervision applies
only to <reflect> and <backtrack> spans.

[USER TEMPLATE]
QUERY
{query_with_every_answer_alias_masked}

TERMINAL OUTCOME
{success_or_failure}

MASKED TRAJECTORY TREE
{active_and_archived_nodes_with_masked_answers}

STRUCTURED HINT
{"verified_state": ..., "first_risk": ..., "return_node": ...,
 "next_decision": "continue|backtrack", "lesson": ...}
\end{tcblisting}
\caption{Answer-masked EMA-teacher context for forced scoring of the student's
existing continuation. Fast-channel supervision covers \reflect{} and
\backtrack{} spans.}
\label{fig:distill_prompt}
\end{figure*}

\begin{figure*}[!t]
\centering
\begin{tcblisting}{promptbox,title={SFT trajectory distillation: wrong-answer intervention}}
[SYSTEM]
Use the private reference only to diagnose why the proposed answer is wrong.
Never quote, identify, paraphrase, or otherwise reveal the reference answer.
The public generation prompt already ends with <think>. Output exactly its
continuation in this form:

concise diagnosis</think><reflect>evidence-based failure assessment</reflect><backtrack>the mistaken branch and a different high-information direction</backtrack>

Do not output an answer, observation, update, Markdown, an opening <think> tag,
or any text after </backtrack>.

[PRIVATE USER TEMPLATE]
QUERY
{query}

FAILED PUBLIC TRACE
{active_trace}

PROPOSED WRONG ANSWER
{proposed_answer}

PRIVATE REFERENCE ANSWER(S)
{reference_aliases}
\end{tcblisting}
\caption{Reference-aware SFT intervention that converts a premature wrong
answer into a diagnosis, backtrack decision, and new search direction.}
\label{fig:sft_prompt}
\end{figure*}

\section{Retrieval and Baseline Controls}
\label{appendix:baselines}

The experiments compare four baseline families from the main text: no-RAG
(Base and CoT), naive-RAG (FS-RAG and FL-RAG), agentic-RAG (ReAct, IRCoT,
TCRAG, and ReSearch), and RL-based agentic-RAG (Search-R1, AEPO, ARPO, Mem1,
and AgenticRAG-R1). The mathematical-reasoning comparison additionally includes
RLSD. Scale-matched experiments use Qwen2.5-3B-Instruct and
Qwen2.5-7B-Instruct.

All retrieval methods share the November 1, 2023 English Wikipedia snapshot,
retriever, top-$k$ setting, context budget, tool-call budget, decoding
configuration, and answer normalizer. We follow the local-search setup of
Search-R1~\citep{jin2025search}. Each \texttt{<search>} action is passed to the
local retriever with $k=3$, and the three retrieved title--passage pairs are
inserted into the active trajectory as an \texttt{<observation>}. The policy
therefore bases subsequent search, reflection, and answer actions on passage
content presented in a common retrieval format.

\section{Reflection Protocol and Answer-Masked Feedback}
\label{appendix:protocol}

\subsection{Action Protocol.}
At each model turn, the policy emits a nonempty \texttt{<think>} span followed
by one public action: \texttt{<search>}, \reflect{}, \reflect{} followed by
\backtrack{}, or \answer{}. The controller appends retrieved observations after
search and compressed \texttt{<update>} messages after backtracking. Policy
loss masks select the model-generated reasoning and action tokens.

For a \backtrack{} request, the controller evaluates preceding checkpoints in
reverse order and resumes from the latest checkpoint that supports the proposed
correction. The root provides a clean-slate recovery point. The abandoned
suffix becomes an archived branch, and the resumed active context contains the
validated prefix plus a compact forward-looking update.

\subsection{Teacher Schema.}
For each sampled control turn, the EMA teacher receives the query, the complete
serialized tree with active and archived markers, and a binary terminal outcome.
Every \answer{} span and every known answer alias is mapped to
\texttt{[MASKED\_ANSWER]} before tokenization. The structured hint contains:

\begin{itemize}
    \item \texttt{verified\_state}: evidence supported by retrieved observations;
    \item \texttt{first\_risk}: the earliest unsupported inference or state contamination;
    \item \texttt{return\_node}: the checkpoint proposed for recovery;
    \item \texttt{next\_decision}: \texttt{continue} or \texttt{backtrack}; and
    \item \texttt{lesson}: a compact forward-looking constraint for working memory.
\end{itemize}

The first four fields instantiate the control label $\mathbf{h}_t$ in the main
text, and \texttt{lesson} materializes the compressed update $u_{j:t}$. The
teacher force-scores the student's existing continuation under this structured
context. The fast token mask covers sampled tokens inside \reflect{} and
\backtrack{} spans. Alias matches set the span mask to zero and replace the
teacher-side span with length-preserving neutral tokens before scoring.

\section{Reversible Trajectory Representation}
\label{appendix:serialization}

Each trajectory-tree node stores its identifier, parent, action type, text or
environment observation, child identifiers, and generation metadata. The tree
uses append-only state transitions. Backtracking archives the abandoned suffix
and creates a new active branch after the selected checkpoint. Subsequent model
generation receives the query, compressed memory, and active path; the EMA
teacher receives the answer-masked global tree for control diagnosis.

A sampled execution can produce several compact training records sharing one
trajectory identifier. Slow-channel masks select policy tokens on the final
active execution. Tool observations, controller updates, protocol suffixes,
and archived-branch tokens use a zero slow mask. Archived \reflect{} and
\backtrack{} spans can retain the fast mask as diagnostic supervision. For GRPO
normalization, compact siblings are collapsed by trajectory identifier and then
grouped by the original prompt identifier, assigning one outcome contribution
to each sampled trajectory.

\section{Two-Channel Training Algorithm}
\label{appendix:algorithm}

Algorithm~\ref{alg:loongreflect_training} summarizes the training procedure
using realized parameter displacements, including Adam and LoRA updates. Let
$\theta_0$ denote the outer parameters, $\widetilde\theta$ the provisional fast
candidate, and $\theta_s$ the candidate after one slow update. Define
$d_f=\widetilde\theta-\theta_0$ and $d_s=\theta_s-\widetilde\theta$. These
descent displacements use the same inner-product conflict test as $g_f$ and
$g_s$ in the main text.

\begin{algorithm}[H]
\caption{Answer-masked two-channel training with look-ahead coordination}
\label{alg:loongreflect_training}
\begin{algorithmic}[1]
\REQUIRE Policy $\pi_\theta$, EMA teacher $q_{\bar\theta}$, $K=3$, $w=1$
\FOR{outer step $t=1,\ldots,100$}
    \STATE Sample 8 prompts and 4 trajectories per prompt
    \STATE Build reversible trees and binary terminal rewards
    \STATE Select final active executions for the slow loss
    \STATE Build answer-masked teacher hints and control-token masks
    \STATE Snapshot trainable parameters $\theta_0$
    \FOR{$k=1,\ldots,K$}
        \STATE Apply one provisional update using $\mathcal{L}_{\mathrm{fast}}$
    \ENDFOR
    \STATE Snapshot $\widetilde\theta$ and rescore the sampled trajectories
    \STATE Apply one provisional update using $\mathcal{L}_{\mathrm{slow}}$
    \STATE Snapshot $\theta_s$; set $d_f\leftarrow\widetilde\theta-\theta_0$
    \STATE Set $d_s\leftarrow\theta_s-\widetilde\theta$
    \IF{$\langle d_f,d_s\rangle<0$}
        \STATE $d_f\leftarrow d_f-\frac{\langle d_f,d_s\rangle}{\|d_s\|_2^2}d_s$
    \ENDIF
    \STATE Commit $\theta\leftarrow\theta_0+d_s+w d_f$
    \STATE Update the EMA teacher after the fused commit
\ENDFOR
\end{algorithmic}
\end{algorithm}

The snapshots and fused displacement cover trainable LoRA parameters. Inner
products and norms are reduced across distributed parameter shards. Optimizer
moments continue across the candidate substeps while the two parameter-space
directions are constructed.

\section{Training Configuration}
\label{appendix:hyperparameters}

Tables~\ref{tab:sft_hyperparameters}--\ref{tab:rl_hyperparameters} report the
SFT, Qwen3-32B trajectory-generation, and two-channel RL configurations
separately. Both model sizes use \textbf{100 outer RL steps}. Each outer step
contains three fast candidate updates followed by one slow candidate update.
RL initializes the policy from the SFT weights and initializes a new optimizer,
RNG stream, and iteration counter for the 100-step run.

\begin{table}[t]
\centering
\fontsize{7.25}{8.15}\selectfont
\setlength{\tabcolsep}{3.5pt}
\renewcommand{\arraystretch}{1.05}
\begin{tabular}{p{0.39\columnwidth}p{0.51\columnwidth}}
\toprule
\rowcolor{gray!10}
\textbf{Parameter} & \textbf{Value} \\
\midrule
Backbones & Qwen2.5-3B / Qwen2.5-7B \\
Training sources & HotpotQA + 2WikiMultiHopQA \\
Retained trajectories & 600 \\
Source proportion & Original combined-pool ratio \\
Epochs & 1 \\
Learning rate & $5\times10^{-6}$ \\
Global batch / sequence length & 64 / 16,384 \\
LoRA rank / scaling & 64 / 128 \\
Optimizer / Adam $\beta_1,\beta_2$ & Adam / 0.9, 0.95 \\
Scheduler / minimum LR / warmup & cosine / $1\times10^{-6}$ / 10\% \\
Weight decay / gradient clip & 0.1 / 1.0 \\
Training precision & BF16 \\
Loss boundary & Complete sampled-continuation tokens \\
Megatron / shuffle seed & 1234 / 42 \\
RL initialization & SFT weights; new optimizer and RNG \\
GPU layout & 1 node / 8 actor GPUs \\
Target unit & Public model turn \\
\bottomrule
\end{tabular}
\caption{Learner SFT configuration.}
\label{tab:sft_hyperparameters}
\end{table}

\begin{table}[t]
\centering
\fontsize{7.25}{8.15}\selectfont
\setlength{\tabcolsep}{3.5pt}
\renewcommand{\arraystretch}{1.05}
\begin{tabular}{p{0.42\columnwidth}p{0.48\columnwidth}}
\toprule
\rowcolor{gray!10}
\textbf{Parameter} & \textbf{Value} \\
\midrule
Teacher & Qwen3-32B (local) \\
Selection filters & success, reflection, length, Search-R1 failure \\
Rejection comparison / top-$k$ & Search-R1 / 3 \\
Tensor parallelism / context & 8 / 16,384 \\
Static memory fraction & 0.82 \\
Attention / execution & FA3 / eager \\
Actor temperature / top-$p$ & 0.6 / 0.95 \\
Trajectory / turn token cap & 16,384 / 1,024 \\
Interaction / minimum turns & 16 / 5 \\
Controller temperature / cap & 0.1 / 768 \\
Teacher retries & 2 \\
Workers / maximum in flight & 32 / 64 \\
LLM concurrency / maximum & 32 / 64 \\
Reference use & verification and recovery diagnosis \\
Control-target screening & normalized reference-alias filter \\
GPU layout & 1 node / tensor parallelism 8 \\
\bottomrule
\end{tabular}
\caption{Qwen3-32B trajectory generation and rejection-sampling configuration.}
\label{tab:teacher_hyperparameters}
\end{table}

\begin{table}[t]
\centering
\fontsize{7.15}{8.0}\selectfont
\setlength{\tabcolsep}{3.2pt}
\renewcommand{\arraystretch}{1.03}
\begin{tabular}{p{0.43\columnwidth}p{0.47\columnwidth}}
\toprule
\rowcolor{gray!10}
\textbf{Parameter} & \textbf{Value} \\
\midrule
Training sources & HotpotQA + 2WikiMultiHopQA \\
Outer steps & 100 \\
Prompts / trajectories per step & 8 / 32 \\
Trajectories per prompt & 4 \\
Global batch size & 32 \\
Temperature / top-$p$ & 1.0 / 1.0 \\
Prompt / response limit & 2,048 / 16,384 \\
Context / sequence length & 16,384 / 16,384 \\
Interaction / turn-token cap & 16 / 1,024 \\
Terminal reward & normalized exact match in $\{0,1\}$ \\
Optimizer / learning rate & Adam / $1\times10^{-6}$ \\
Adam $\beta_1,\beta_2$ & 0.9 / 0.98 \\
Scheduler / warmup & constant / 0 \\
Gradient clip / precision & 1.0 / BF16 \\
Weight decay & 0.01 \\
GRPO clip low / high & 0.20 / 0.28 \\
Policy KL coefficient & 0.001 \\
Reverse-KL clip $c$ & 10 \\
EMA decay & cosine 0.996 $\rightarrow$ 1.0 \\
Fast updates $K$ / weight $w$ & 3 / 1 \\
Action protocol & search, reflect, backtrack, answer \\
LoRA rank / scaling & 64 / 128 \\
Retrieval snapshot & English Wikipedia / 2023-11-01 \\
Search-R1 top-$k$ / passages & 3 / 3 \\
Policy observation & title and passage content \\
Megatron / shuffle seed & 1234 / 42 \\
Actor / rollout GPU layout & 1 node: 4 / 4 \\
Actor tensor parallelism (3B / 7B) & 2 / 4 \\
\bottomrule
\end{tabular}
\caption{Two-channel RL configuration. The horizon contains 100 outer steps for
both model sizes.}
\label{tab:rl_hyperparameters}
\label{tab:training_hyperparameters}
\end{table}

\subsection{Hyperparameter Selection.}
The sensitivity study varies $K\in\{1,2,3,4\}$ with $w=1$ and
$w\in\{0.5,1,2\}$ with $K=3$. The selected setting $K=3,w=1$ achieves the
highest average QA F1 in these comparisons. All remaining hyperparameters use
the single prespecified values in Tables~\ref{tab:sft_hyperparameters}--
\ref{tab:rl_hyperparameters} throughout the evaluation datasets.

\subsection{Compute and Software Environment.}
The 3B and 7B training runs use one Ubuntu 22.04.5 node with eight NVIDIA
A800-SXM4-80GB GPUs, two Intel Xeon Platinum 8358P CPUs, and approximately
2 TiB of system memory. The software stack consists of Python 3.12.13, PyTorch
2.11.0+cu129, Ray 2.56.0, SGLang 0.5.12.post1, Transformers 5.6.0, Slime
0.3.0, and Megatron Core 0.16.0rc0. The upstream commit prefixes are
\texttt{90c212b} for Slime, \texttt{1dcf0da} for Megatron-LM, and
\texttt{5a15cde} for SGLang.

\section{Evaluation and Reporting Protocol}
\label{appendix:evaluation}

For the seven QA datasets and two mathematical-reasoning datasets, we report
answer-level token F1 as a percentage. The normalizer lowercases text, removes
punctuation and English articles, and collapses whitespace. Precision and
recall are computed from token overlap, and multiple references use the
maximum reference F1. This metric credits partially matching short-answer
aliases across heterogeneous benchmarks. Training uses normalized exact match
as the terminal reward, and evaluation computes F1 from frozen checkpoints.

Megatron actors use seed 1234 for Python, NumPy, PyTorch, and model-parallel
CUDA RNG initialization. Rollout dataset shuffling uses seed 42. SGLang server
seeds are derived by rank from the Megatron seed. Reported tables contain point
estimates from this recorded seed configuration.

\section{Long Reflective Case Study}
\label{appendix:case}

Figure~\ref{fig:long_case} provides a long reflective case from the
2WikiMultiHopQA development set. The question asks for the death date of Joanna
of Aragon's mother, and the trajectory recovers from a spouse-substitution
error before producing the final answer.

\begin{figure*}[!t]
\centering
\begin{tcolorbox}[
  enhanced,
  colback=gray!7,
  colframe=gray!32,
  boxrule=0.45pt,
  arc=2mm,
  outer arc=2mm,
  left=2mm,right=2mm,top=1.2mm,bottom=1.2mm
]
\fontsize{7.2}{8.1}\selectfont
\textbf{Question (2WikiMultiHopQA dev):}
When did Joanna Of Aragon, Countess Of Foix's mother die?
\hfill
\textbf{Reference:} 23 October 1378
\end{tcolorbox}

\noindent
\begin{minipage}[t]{0.488\textwidth}
\raggedright
\textbf{\small Initial branch and diagnosis}

\begin{tcolorbox}[thinkaction,title={\texttt{1.\ <think>}}]
The question contains two linked facts: the identity of Joanna's mother and
that woman's exact death date. I should not search for a date until the family
relation is anchored, because John I of Aragon had more than one wife and a
spouse biography could easily be attached to the wrong maternal branch. I will
therefore retrieve Joanna's parentage first and preserve the named mother as
the entity constraint for the second hop.
\end{tcolorbox}
\begin{tcolorbox}[searchaction,title={\texttt{2.\ <search>}}]
Joanna of Aragon Countess of Foix mother parents
\end{tcolorbox}
\begin{tcolorbox}[observationaction,title={\texttt{3.\ <observation>}}]
\textbf{[1]} Joanna of Aragon (1375--1407), Countess of Foix, was a daughter
of John I of Aragon and his first wife, Martha of Armagnac.  \textbf{[2]}
Martha of Armagnac was John I's first wife; accounts of their children include
Joanna, who later became Countess of Foix.  \textbf{[3]} John I first married
Martha of Armagnac and later married Yolande of Bar, placing Joanna in the
children of the first marriage.
\end{tcolorbox}
\begin{tcolorbox}[thinkaction,title={\texttt{4.\ <think>}}]
All three passages favor Martha, but the third result introduces Yolande as a
second spouse with a separate biography. A broad search about John I's wife
could rank Yolande highly and tempt me to use her date merely because it is
easy to retrieve. I will inspect that competing spouse record to test whether
it belongs to Joanna's branch; this creates a deliberate alternative branch
whose relation must be checked before any date is accepted.
\end{tcolorbox}
\begin{tcolorbox}[searchaction,title={\texttt{5.\ <search>}}]
Yolande of Bar death date wife of John I of Aragon Joanna mother
\end{tcolorbox}
\begin{tcolorbox}[observationaction,title={\texttt{6.\ <observation>}}]
\textbf{[1]} Yolande of Bar was the second wife of John I of Aragon and died
on 3 July 1431.  \textbf{[2]} John I's first wife was Martha of Armagnac;
after Martha's death he married Yolande, who became queen consort of Aragon.
\textbf{[3]} Joanna of Aragon belonged to John I and Martha's first-marriage
branch, whereas Yolande's children form the later-marriage branch.
\end{tcolorbox}
\begin{tcolorbox}[reflectaction,title={\texttt{7.\ <reflect>}}]
\textbf{Verified state:} Martha is Joanna's mother, while Yolande is John I's
later wife; the retrieved date 3 July 1431 is explicitly Yolande's.
\textbf{Diagnosis:} the branch drifted from the required relation
``Joanna's mother'' to the looser relation ``a wife of Joanna's father.'' The
date is locally supported yet answers the wrong entity. \textbf{Return node:}
the first observation, where Joanna is directly linked to Martha.
\textbf{Recovery decision:} discard the spouse-substitution branch and issue
an entity-specific query for Martha's full dates.
\end{tcolorbox}
\begin{tcolorbox}[backtrackaction,title={\texttt{8.\ <backtrack>}}]
I return to the last checkpoint supported by the parent relation: Joanna is
the daughter of Martha of Armagnac. The Yolande evidence remains useful only
as an explanation of the ambiguity and contributes no candidate answer. From
this checkpoint onward, every date must be attached explicitly to Martha
before it can update the answer state.
\end{tcolorbox}
\end{minipage}
\hfill
\begin{minipage}[t]{0.488\textwidth}
\raggedright
\textbf{\small Recovered branch and verified answer}

\begin{tcolorbox}[updateaction,title={\texttt{9.\ <update>}}]
\textbf{Compressed verified state:} Joanna of Aragon, Countess of Foix, is a
daughter of John I of Aragon and his first wife, Martha of Armagnac; therefore
the target entity is Martha. Yolande of Bar is the second wife and is excluded
from the answer path. The remaining task is to retrieve Martha's exact death
date and then reconnect that date to the preserved Joanna--Martha relation.
\end{tcolorbox}
\begin{tcolorbox}[thinkaction,title={\texttt{10.\ <think>}}]
The recovered state fixes the person, so the next query should name Martha
directly and request both endpoints of her lifespan. Full biographical dates
are preferable to a query containing a guessed answer, since they provide an
independent passage and reduce confirmation bias. I will compare all three
returned passages for agreement on the day, month, and year.
\end{tcolorbox}
\begin{tcolorbox}[searchaction,title={\texttt{11.\ <search>}}]
Martha of Armagnac birth and exact death date
\end{tcolorbox}
\begin{tcolorbox}[observationaction,title={\texttt{12.\ <observation>}}]
\textbf{[1]} Martha of Armagnac (18 February 1347--23 October 1378) was the
first wife of John I of Aragon.  \textbf{[2]} The marriage of John and Martha
continued until Martha died in Zaragoza on 23 October 1378; John subsequently
married Yolande of Bar.  \textbf{[3]} A genealogical account lists Martha's
death as 23 October 1378 and identifies her as the mother of children from
John I's first marriage.
\end{tcolorbox}
\begin{tcolorbox}[reflectaction,title={\texttt{13.\ <reflect>}}]
\textbf{Progress check:} the top three passages independently converge on
23 October 1378, and each attaches the date to Martha rather than Yolande.
\textbf{Evidence gap:} the death-date retrieval and the initial parentage
retrieval are concordant but occur in separate turns. \textbf{Risk control:}
before answering, I should retrieve one final result set that names Joanna and
Martha together, so that the two-hop chain is explicit in the active branch.
\textbf{Next decision:} query the relationship directly and use the returned
passages as a cross-check rather than opening another spouse branch.
\end{tcolorbox}
\begin{tcolorbox}[thinkaction,title={\texttt{14.\ <think>}}]
The answer candidate is stable, but a defensible multi-hop answer should show
the join, not only two isolated facts. I will include Joanna's title, Martha's
name, and the maternal relation in the same query. If the top three passages
preserve that identity and at least one also repeats the date, the chain will
support the exact response without relying on the discarded Yolande branch.
\end{tcolorbox}
\begin{tcolorbox}[searchaction,title={\texttt{15.\ <search>}}]
Joanna of Aragon Countess of Foix mother Martha of Armagnac
\end{tcolorbox}
\begin{tcolorbox}[observationaction,title={\texttt{16.\ <observation>}}]
\textbf{[1]} Joanna, Countess of Foix, was a daughter of John I of Aragon and
Martha of Armagnac.  \textbf{[2]} Martha's children with John I included
Joanna of Aragon, and Martha died on 23 October 1378.  \textbf{[3]} A royal
family record connects Joanna to John I's first marriage with Martha and gives
Martha's lifespan as 18 February 1347 to 23 October 1378.
\end{tcolorbox}
\begin{tcolorbox}[reflectaction,title={\texttt{17.\ <reflect>}}]
\textbf{Final verification:} the active branch now contains the complete
chain Joanna $\rightarrow$ mother Martha $\rightarrow$ death on 23 October
1378. The final top-three set repeats both the relationship and the date, while
the earlier date-focused set supplies independent agreement. \textbf{Resolved
risk:} Yolande's 1431 date is confined to the archived spouse-substitution
branch and cannot contaminate the answer. \textbf{Answer decision:} return the
exact date in a concise form matching the requested temporal value.
\end{tcolorbox}
\begin{tcolorbox}[answeraction,title={\texttt{18.\ <answer>}}]
\textbf{October 23, 1378}
\end{tcolorbox}
\end{minipage}
\caption{Long reflective case from 2WikiMultiHopQA. Each observation displays
the three passages returned by Search-R1 local retrieval ($k=3$). Purple
denotes reasoning, green search, gray retrieved observations, orange
reflection, red backtracking, blue compressed updates, and yellow the final
answer.}
\label{fig:long_case}
\end{figure*}

\FloatBarrier
\section{Limitations and Ethical Considerations}
\label{appendix:limitations}

The method depends on retrieval quality and on the controller's checkpoint and
recovery judgments. Retrieval noise can affect verified-state construction,
checkpoint selection, and the compressed update supplied to the next branch.
Errors in an early diagnosis can therefore propagate through later search and
answer decisions. Normalized exact-match reward can also under-credit valid
semantic aliases that receive partial credit under token F1. The EMA teacher's
binary terminal outcome provides useful trajectory-level context and may shape
the style of local diagnoses.

Reversible multi-turn search and the three-fast/one-slow look-ahead update add
inference and optimization cost. The current experiments focus on two training
datasets, seven QA evaluations, two mathematical-reasoning evaluations, and
the recorded seed configuration. Future work can broaden the task domains,
compare multiple independent seeds, quantify robustness to retrieval noise,
measure checkpoint-selection accuracy, and analyze error propagation through
compressed updates. A wider study of coordination schedules can also examine
adaptive choices of $K$ and $w$.

The experiments use established public QA and mathematical-reasoning
benchmarks together with an English Wikipedia retrieval corpus. Their topical,
linguistic, and geographic coverage shapes the learned retrieval and reasoning
behavior. Responsible use should follow the licenses of the datasets,
Wikipedia snapshot, and model checkpoints and should evaluate factuality in
the intended deployment domain.

\section{The Use of Large Language Models}
\label{appendix:llm_use}

In this work, Large Language Models (LLMs) supported language polishing and
programming tasks, including improvements to grammar, clarity, and readability,
as well as general coding suggestions and debugging assistance. The authors
carefully reviewed and verified all LLM-assisted outputs and retain full
responsibility for the study's conceptualization, experimental design, result
analysis, and conclusions.


\end{document}